\documentclass[]{sensenova}

\usepackage[utf8]{inputenc}
\usepackage[T1]{fontenc}
\usepackage{url}
\usepackage{amsmath,amssymb,bm}
\usepackage{booktabs,multirow,tabularx,array,makecell}
\usepackage{graphicx}
\usepackage{subcaption}
\usepackage[table]{xcolor}
\usepackage{wrapfig}
\usepackage{tikz}
\usetikzlibrary{
  positioning,
  arrows.meta,
  fit,
  calc,
  shapes.geometric,
  backgrounds
}
\usepackage{float}
\usepackage{placeins}
\usepackage{siunitx}
\usepackage{pifont}
\usepackage{fontawesome5}
\usepackage{pgfplots}
\pgfplotsset{compat=1.18}
\usepgfplotslibrary{groupplots}
\usetikzlibrary{matrix,decorations.pathreplacing}
\usepackage{longtable}

\AtEndPreamble{
  \crefname{figure}{Figure}{Figures}
  \crefname{section}{Section}{Sections}
  \crefname{table}{Table}{Tables}
  \crefname{equation}{Eq.}{Eqs.}
  \crefname{appendix}{Appendix}{Appendices}
  \Crefname{figure}{Figure}{Figures}
  \Crefname{section}{Section}{Sections}
  \Crefname{table}{Table}{Tables}
  \Crefname{equation}{Eq.}{Eqs.}
  \Crefname{appendix}{Appendix}{Appendices}
}

\definecolor{themebrown}{HTML}{A1846F}
\definecolor{themeink}{HTML}{8A6D5A}
\colorlet{sensepurple}{themebrown}
\renewcommand{\sensepurple}[1]{{\bfseries\color{themebrown} #1}}

\hypersetup{
  linkcolor=themeink,
  citecolor=themeink,
  urlcolor=themeink,
  pdftitle={Probing Spatial State Representations in Visual Foundation Models},
  pdfauthor={Baiqiao Yin, Zhongang Cai, Ruisi Wang, Qiwei Liang, Lei Yang, Ziwei Liu}
}

\definecolor{accentorange}{HTML}{926860}
\definecolor{paleorange}{HTML}{EEE0D8}
\definecolor{warmwhite}{HTML}{DCD2C2}
\definecolor{warmgray}{HTML}{E8E4DE}
\definecolor{graybrown}{HTML}{4E4842}
\definecolor{structblue}{HTML}{DCE8EC}
\definecolor{memorygreen}{HTML}{E4EAD9}
\definecolor{morandibest}{HTML}{B0C0A8}
\definecolor{morandirule}{HTML}{A8A096}
\definecolor{headergray}{HTML}{F1F1EE}
\definecolor{morandisecond}{HTML}{CEB8A0}
\definecolor{morandislate}{HTML}{5C6C70}
\definecolor{morandisage}{HTML}{566858}
\definecolor{structink}{HTML}{326F83}
\definecolor{memoryink}{HTML}{687E3E}
\definecolor{predictink}{HTML}{A65F38}
\colorlet{morandiink}{graybrown}
\colorlet{morandigray}{warmgray}
\colorlet{predictorange}{paleorange}
\colorlet{morandisand}{warmwhite}
\colorlet{morandiclay}{accentorange}

\colorlet{morandiheader}{headergray}
\colorlet{morandiblue}{structblue}
\colorlet{morandigreen}{memorygreen}
\colorlet{morandirose}{predictorange}
\colorlet{softgray}{morandigray}
\colorlet{softblue}{morandiblue}
\colorlet{softgreen}{morandigreen}
\colorlet{softorange}{morandirose}
\colorlet{draftred}{morandiclay}
\colorlet{draftblue}{morandislate}

\arrayrulecolor{morandirule}

\newcommand{\best}[1]{\cellcolor{morandibest}\textbf{#1}}

\newcommand{\negnum}[1]{$-$#1}

\newcommand{\model}[1]{\textsc{#1}}
\newcommand{\levelname}[1]{\textbf{L#1}}

\newcommand{\taskgeometry}{Metric geometry}

\newtcolorbox{keyfinding}[2]{%
  enhanced,
  breakable,
  colback=#1!60!white,
  colframe=#1!45!graybrown,
  coltext=graybrown,
  boxrule=0pt,
  borderline west={2.5pt}{0pt}{#1!60!graybrown},
  arc=1.5pt,
  left=9pt,
  right=9pt,
  top=7pt,
  bottom=7pt,
  before skip=9pt,
  after skip=9pt,
  fontupper=\normalsize,
  before upper={%
    {\normalsize\bfseries #2}\par
    \vspace{3pt}%
  },
}

\newtcbox{\linkpill}{on line,
  colback=themebrown!8, colframe=themebrown!60, boxrule=0.5pt, arc=6pt,
  boxsep=0pt, left=7pt, right=7pt, top=3pt, bottom=3pt,
  fontupper=\small\color{themeink}}
\newcommand{\paperlink}[3]{\href{#1}{\linkpill{#2\hspace{0.45em}#3}}}
\newcommand{\paperlinks}{%
  \paperlink{https://yyyybq.github.io/probing-spatial-state/}%
    {\faGlobe}{Project Page}%
  \hspace{0.8em}%
  \paperlink{https://github.com/yyyybq/Probing_Spatial_State_Release}%
    {\faGithub}{Code}}

\newcommand{\reportlogo}{assets/affiliation_logos.png}
\fancypagestyle{firststyle}{%
  \fancyhf{}%
  \fancyhead[L]{%
    \vskip 2mm
    \IfFileExists{\reportlogo}{\includegraphics[height=10mm]{\reportlogo}}{}%
  }%
  \fancyfoot[C]{\thepage}%
}

\title{Probing Spatial State Representations\\in Visual Foundation Models}

\author[1,2]{Baiqiao Yin}
\author[2]{Zhongang Cai}
\author[2]{Ruisi Wang}
\author[3]{Qiwei Liang}
\author[4]{Lei Yang}
\author[2]{Ziwei Liu}

\affiliation[1]{New York University}
\affiliation[2]{Nanyang Technological University}
\affiliation[3]{The Hong Kong University of Science and Technology}
\affiliation[4]{The Chinese University of Hong Kong}

\contribution{\protect\paperlinks}

\abstract{
An observer sees only part of the surrounding world, yet must keep track
of what lies beyond the current view. Spatial perception therefore
involves more than recognizing individual images: it calls for a
persistent spatial state that is updated under self-motion and used to
guide action. Do visual foundation models provide representations that
support such a state?
We study this question through four levels of spatial capability---spatial
structure, object persistence, predictive spatial update, and
actionability---using frozen-feature probes on ten representations under
streaming egocentric observations of static indoor environments.
The resulting capability profiles vary across tasks, network layers, and
prefix lengths: strong readout of the visible scene does not ensure
accurate object persistence or future-feature retrieval.
Matched controls separate readout accuracy from reliance on visual
history and motion. Native future prediction retains recoverable spatial
structure, but a spatial reader trained on real features does not transfer
to predicted features unchanged.
Finally, goal-dependent candidate selection does not establish reliable
continuous action utility: predicted actions respond to goal changes, yet
supplying the correct goal does not reliably improve progress.
Across the four levels, object persistence and actionability build on
camera-pose readout, whereas feature prediction builds on visible-scene
readout.
Together, these findings motivate evaluating spatial representations not
only by what can be decoded from them, but by whether that information is
maintained beyond observation, updated predictively, and used for
goal-directed action.
}

\checkdata[Keywords]{visual foundation models, spatial state
representations, probing, object persistence, actionability.}

\begin{document}
\raggedbottom

\maketitle

\begin{figure}[t]
  \centering
  \includegraphics[width=0.9\textwidth]{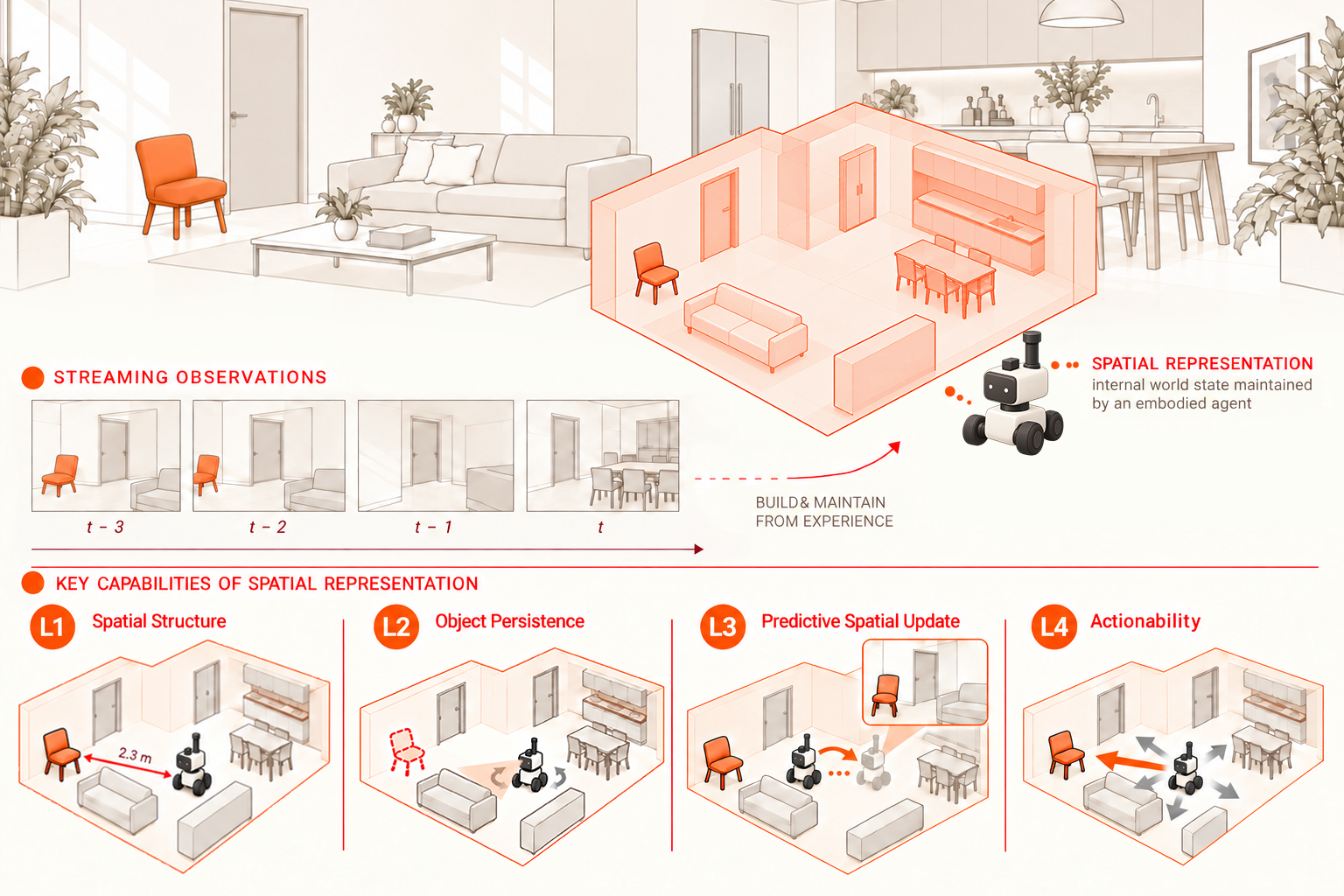}
  \caption{\textbf{Spatial state representations from streaming egocentric observations.}
  A moving observer builds an internal spatial representation from partial
  views of an environment.
  \textbf{L1} (\emph{spatial structure}) captures scene geometry and object
  identity across views.
  \textbf{L2} (\emph{object persistence}) retains the observer-relative
  locations of previously seen objects after they leave the field of view.
  \textbf{L3} (\emph{predictive spatial update}) anticipates what the
  observer would see after an action or a sequence of actions.
  \textbf{L4} (\emph{actionability}) uses the represented state and a
  specified goal to select an action that makes progress toward that goal.}
  \label{fig:teasor}
\end{figure}

\section{Introduction}
\label{sec:introduction}

Spatial representation is one of the most fundamental capacities of
biological perception. Humans and other animals do not experience the world
as a sequence of independent retinal images. Through continuous exploration,
they maintain information about the surrounding space, keep track of objects
that are no longer visible, update their own location as they move, and use
this information to guide subsequent behavior. These abilities are not
inherently linguistic: they arise from perception and from sensorimotor
interaction with the environment.

Several classical ideas in perception and spatial cognition point to this
view. Object-file theories describe how object-specific information remains
associated with the same entity across changes in its visual appearance
\citep{KAHNEMAN1992175}. Spatial updating and path integration describe how
an internal estimate of location is continuously revised from self-motion
\citep{McNaughton2006}. Sensorimotor contingency theories further emphasize
that perception involves knowledge of how sensory input would change under
possible actions \citep{oregan2001}. Together, these perspectives suggest
that a useful representation of space should describe more than what is
visible in the current image: it should maintain a state of the surrounding
world that persists and evolves as the observer interacts with it.

An observer's \textbf{spatial state} is the configuration of the surrounding
environment relative to the observer, including previously observed objects
that are no longer visible. A \textbf{spatial state representation} is an
internal encoding of this configuration inferred from the observer's visual
history. We do not prescribe an explicit 3D map, an object-centric scene
graph, a coordinate system, or a neural architecture. Instead, we
characterize the functional usefulness of a representation through four
capabilities:

\begin{enumerate}
  \item[\textbf{L1}] \textbf{Spatial structure.}
  Recover the spatial organization of the observed scene through metric
  geometry, instance identity, and view consistency.
  \item[\textbf{L2}] \textbf{Object persistence.}
  Retain the location of a previously observed object after it has left
  the field of view.
  \item[\textbf{L3}] \textbf{Predictive spatial update.}
  Anticipate changes in the observer-relative state under ego-motion,
  evaluated through direct prediction and path integration.
  \item[\textbf{L4}] \textbf{Actionability.}
  Use the represented state and a specified goal to predict an action that
  makes progress toward that goal.
\end{enumerate}

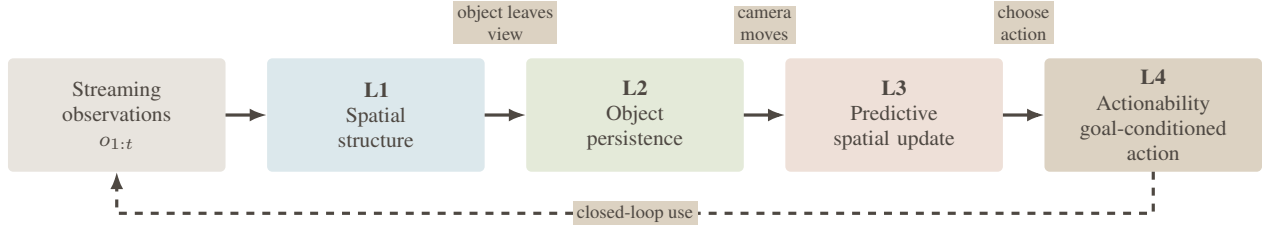
\begin{figure}[t]
  \centering
\begingroup

\pgfmathsetlengthmacro{\spatialboxwidth}
  {(\linewidth - 4*0.55cm)/5 - 8pt}

\begin{tikzpicture}[
  node distance=0.55cm,
  box/.style={
    rounded corners=2pt,
    draw=none,
    minimum height=1.50cm,
    text width=\spatialboxwidth,
    inner xsep=4pt,
    inner ysep=3pt,
    outer sep=0pt,
    align=center,
    font=\footnotesize,
    text=morandiink
  },
  arr/.style={
    -{Latex[length=2.3mm]},
    very thick,
    draw=morandiink
  },
  lab/.style={
    font=\scriptsize,
    align=center,
    fill=warmwhite,
    inner sep=1.5pt,
    text=morandiink
  }
]


  \node[box, fill=morandigray] (obs)
    {Streaming\\observations\\$o_{1:t}$};

  \node[box, fill=morandiblue, right=of obs] (l1)
    {\levelname{1}\\Spatial\\structure};

  \node[box, fill=morandigreen, right=of l1] (l2)
    {\levelname{2}\\Object\\persistence};

  \node[box, fill=morandirose, right=of l2] (l3)
    {\levelname{3}\\Predictive\\spatial update};

  \node[box, fill=morandisand, right=of l3] (l4)
    {\levelname{4}\\Actionability\\goal-conditioned\\action};


  \draw[arr]
    (obs.east) -- (l1.west);

  \draw[arr]
    (l1.east) --
    node[lab, midway, above=0.92cm] {object leaves\\view}
    (l2.west);

  \draw[arr]
    (l2.east) --
    node[lab, midway, above=0.92cm] {camera\\moves}
    (l3.west);

  \draw[arr]
    (l3.east) --
    node[lab, midway, above=0.92cm] {choose\\action}
    (l4.west);


  \draw[arr, dashed]
    (l4.south)
    -- ++(0,-0.55cm) coordinate (loopright)
    -- (loopright -| obs.south) coordinate (loopleft)
    -- (obs.south);

  \node[lab]
    at ($(loopleft)!0.5!(loopright)$)
    {closed-loop use};

\end{tikzpicture}

\endgroup
  \caption{\textbf{Functional relationships among the four spatial capabilities.}
  The levels are ordered by what each demands beyond the current
  observation. L1 grounds the spatial state in observed scene structure.
  L2 extends the demand from representing what is visible to retaining
  information after objects leave the field of view, and L3 requires
  anticipating how the observer-relative state changes under ego-motion.
  L4 connects the represented state to goal-directed action.
  The dashed return arrow illustrates potential closed-loop use: actions
  change the observer's viewpoint and hence the subsequent observations,
  while the environment itself remains static.}
  \label{fig:spatial-state-ladder}
\end{figure}

These capabilities, summarized in \cref{fig:spatial-state-ladder}, are
ordered by what each demands beyond the current observation: L1 reads the
visible scene; L2 additionally requires information retained from earlier
views; L3 requires anticipating views that have not yet been observed; and
L4 requires relating the represented state to a goal and an action. Which
lower-level ingredients the higher levels actually draw on is an empirical
question that we return to in \cref{sec:level-relations}.

We study these functions in a streaming setting: the observer receives past
and current visual evidence but no future observations, and representations
are extracted from causal visual prefixes of increasing length. Each prefix
is encoded independently, so the question is what the representation of a
causal prefix affords, not whether the model maintains an online state.
Our experimental scope is observer motion through static indoor
environments; object dynamics, contact physics, and complete physical world
modeling are outside it.

The question spans the full range of \emph{visual foundation models},
including image and video encoders, 3D models, multimodal models, video
generators, and world models. Strong performance on a model's training
objective does not establish all four spatial functions, and success at one
level need not carry over to the next.

We therefore conduct a \textbf{functional probing study} of ten frozen
representations. Six main diagnostics yield seven measurements of spatial
structure, object persistence, and predictive spatial update, since metric
geometry contributes separate depth and camera-pose metrics; goal-conditioned
action prediction additionally tests actionability in a candidate-selection
and a continuous-output setting. Rather than relying on a single extraction
point, we organize the analyses along two axes, network depth and streaming
context, and add matched controls that alter motion, history, transformation
order, or goals while keeping the relevant predictor and target fixed.

\paragraph{Findings.}
No representation leads on all tasks: Fast3R has the strongest camera-pose
and object-persistence readouts, DINOv2 leads direct future-feature
prediction, and V-JEPA2 leads path integration (\cref{sec:main-benchmark}).
VideoMAE localizes previously seen objects more accurately than DINOv2 and
Qwen2.5-VL despite weaker depth, instance, and view-consistency readouts;
across the ten representations, object persistence aligns with camera-pose
readout rather than with these visible-scene metrics. The profiles shift
with layer, and longer prefixes worsen persistence for all eight models in
the endpoint comparison while their effects on prediction vary
(\cref{sec:layer-prefix-analysis}). Separate matched controls distinguish
\emph{decodability} from \emph{input use}: copying the last feature exceeds
learned prediction for V-JEPA2, and visual history helps V-JEPA2 and Wan
but shows no clear benefit for VideoMAE (\cref{sec:probe-validity}). Native
prediction keeps recoverable geometry but not the original readout
interface (\cref{sec:predictive-spatial-interface}), and input use is not
\emph{functional utility}: changing the goal rotates predicted actions by
30--46$^{\circ}$, yet the correct goal does not reliably improve progress
(\cref{sec:action-output}). Across levels, the results reveal two
ingredient chains: object persistence and actionability build on
camera-pose readout, whereas feature prediction builds on visible-scene
readout (\cref{sec:level-relations}).

\paragraph{Contributions.}
We (i) characterize spatial state representations operationally through
four functional capabilities; (ii) develop a frozen-feature probing
framework covering ten representations, with controls for history, motion,
predictive readout, and goal use; and (iii) show that readout accuracy,
input use, and goal-specific utility dissociate across models, layers, and
prefix lengths, motivating multidimensional evaluation of spatial state.

\section{Probing Spatial State Representations}
\label{sec:formal-framework}

We ask what spatial information is accessible in frozen visual features,
which inputs support that information, and whether it remains useful for
prediction and goal-directed action.

\subsection{Four functional capabilities}
\begin{figure}[t]
  \centering
  \includegraphics[width=\textwidth]{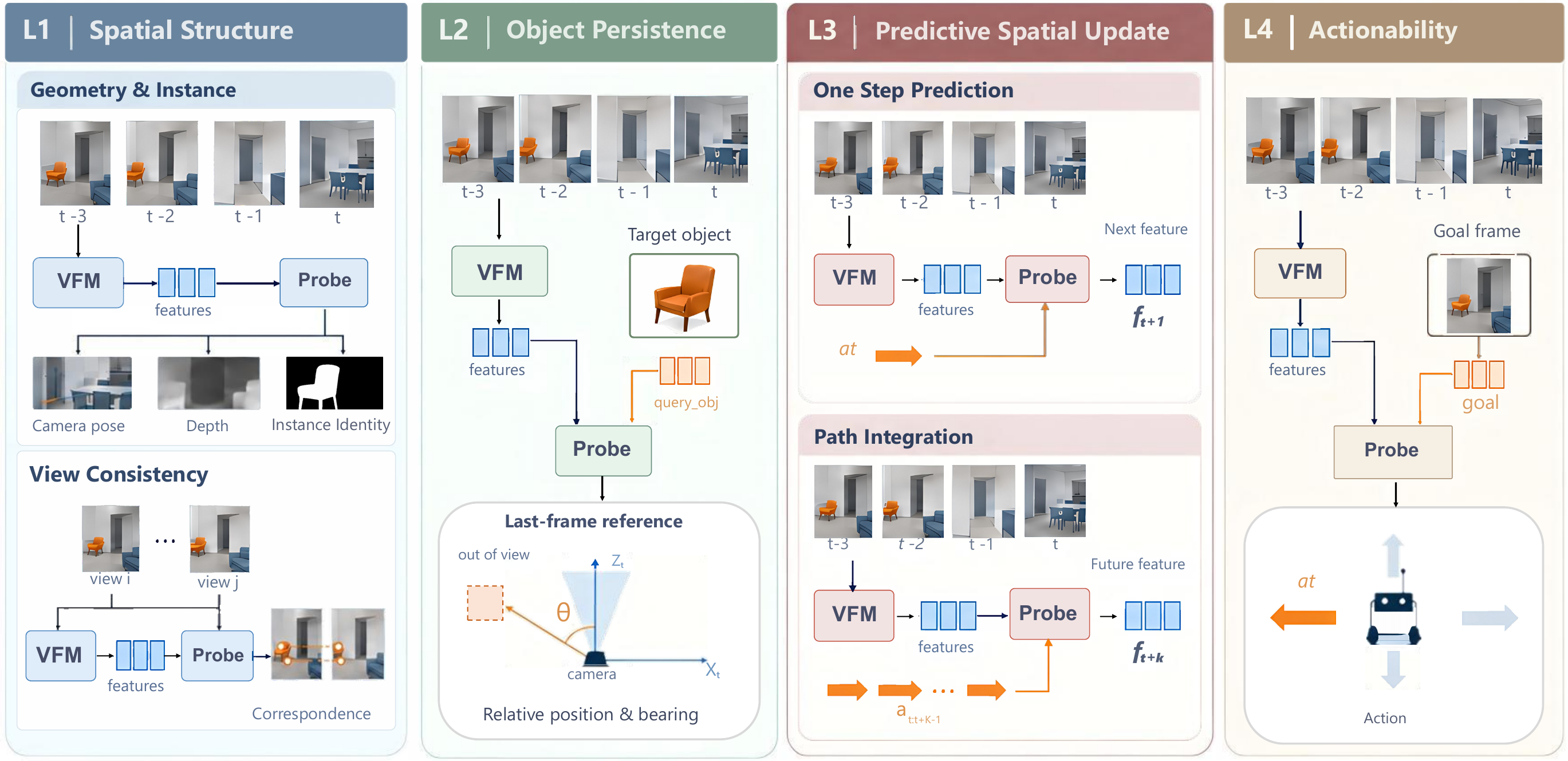}
  \caption{\textbf{Probing four spatial capabilities.} Probe inputs and
  targets at each level: prefix features alone (L1), plus an object query
  (L2), ego-motions and an isolated future feature (L3), or a goal frame
  and an action (L4).}
  \label{fig:taxonomy}
\end{figure}

\Cref{fig:taxonomy} summarizes the probing tasks that instantiate the four
levels of spatial capability.

\paragraph{L1: Spatial structure.}
We first ask what spatial structure is directly readable from the streaming
visual representation. The \textbf{metric geometry} task takes frozen
features of an egocentric video prefix and predicts the current camera pose
and scene depth, testing whether metric 3D structure is accessible. The
\textbf{instance identity} task uses the same features to separate
individual object instances, testing whether the representation organizes
the visible scene at the object level rather than encoding global geometry
alone. The \textbf{view consistency} task matches correspondences between
features extracted from two viewpoints of the same scene, testing whether
spatial correspondences survive changes in viewpoint.

\paragraph{L2: Object persistence.}
We then test whether information about an object remains spatially
available after the object is no longer visible, through
\textbf{out-of-view object localization}. A target object is first observed
in an earlier frame and represented by an object query $q_{\mathrm{obj}}$.
By the end of the video prefix, camera motion has moved the target outside
the observer's field of view. Given the frozen representation of the whole
prefix together with $q_{\mathrm{obj}}$, the probe predicts the target's
location relative to the final camera, evaluated through its bearing. The
task therefore asks whether the representation maintains the spatial state
of a previously observed object beyond direct visual evidence, rather than
merely detecting or recognizing objects that remain visible.

\paragraph{L3: Predictive spatial update.}
Next, we ask whether the representation supports motion-conditioned
prediction of the observer-relative state. In \textbf{direct prediction},
the probe receives the visual prefix and the relative camera motion from
its endpoint to a future observation, and predicts that observation's
feature in a single update. The main evaluation pools targets at
observation offsets $h\in\{1,2,4\}$; MiniMax-H3 uses a separately trained
next-observation-only probe ($h=1$), which we mark explicitly. In
\textbf{path integration}, a recurrent probe instead applies successive
relative motions along the same trajectory and is evaluated at the
waypoints at offsets 1, 2, and 4. The two tasks test direct endpoint
prediction and sequential motion integration, respectively; because their
readout architectures differ, the comparison does not isolate recurrence
alone.

\paragraph{L4: Actionability.}
Finally, we test whether the representation can be used for goal-directed
action through \textbf{goal-conditioned action prediction}. The probe
receives the representation of the current visual history together with an
independently encoded goal frame $q_{\mathrm{goal}}$, and predicts an
action $a_t$ intended to move the observer toward the goal. This level asks
whether the spatial information exposed by the representation is not only
descriptive or predictive, but also usable for selecting behavior
conditioned on a desired future state.

\subsection{Causal streaming formulation}
\label{sec:streaming-evaluation}
For model $m$ and layer $\ell$, the frozen encoder $E^m_{\ell}$ encodes
prefixes that share a start frame and differ only in their endpoint $p$:
\begin{equation}
  \mathbf{Z}^{m,\ell}_{p}
  =E^m_{\ell}\!\left(\mathbf{I}_{1:p}\right),
  \qquad p\in\{8,12,16,24\}.
  \label{eq:causal-streaming-feature}
\end{equation}
Each prefix is encoded independently, contains no frame after $p$, and is
treated as a \emph{candidate spatial state representation} at every level.
For L2, the object query is pooled from frames in which the object was
visible, and the target is its bearing in the camera frame at $p$; neither
the visibility mask nor the target location is given to the probe. No
recurrent state update is required, and because a longer prefix also moves
the endpoint, comparisons across $p$ do not isolate added history.
L3 and L4 additionally use a future frame encoded without access to the
prefix or to any later frame,
\begin{equation}
  \mathbf{Z}^{m,\ell,\mathrm{tar}}_{p+h}
  =E^m_{\ell}\!\left(\mathbf{I}_{p+h}\right),
  \qquad h\in\{1,2,4\},
  \label{eq:isolated-target-feature}
\end{equation}
which serves as the retrieval target for direct prediction and path
integration and as the goal frame for L4. MiniMax-H3's direct prediction is
restricted to $h=1$; the other direct probes and all path probes use
$h\in\{1,2,4\}$. Stochastic encoders use matched noise seeds; further
constructions are specified with each experiment.

\section{Experimental Setup}
\label{sec:experimental-setup}

\subsection{Data and splits}
\label{sec:data-splits}

All temporal experiments use posed ScanNet++ trajectories
\citep{yeshwanth2023scannetpp} processed by our pipeline. Each sample
provides RGB observations, camera poses, instance geometry, and visibility
annotations. Training and validation scenes are disjoint, and scene-level
partitions keep related trajectories, objects, and windows together.
Fast3R's training data include ScanNet++ \citep{yang2025fast3r}, and some
of our validation scenes are among its training scenes. Its leading
camera-pose, object-persistence, and action results may therefore partly
reflect familiarity with these scenes.
Task-specific splits, sample counts, and seeds are specified with each
experiment and in \cref{sec:protocol-details}. Views without a meaningful
temporal trajectory are used only for tasks that require neither history
nor motion.

\subsection{Models and extraction points}
\label{sec:model-layer-grid}

\begin{wraptable}{r}{0.48\columnwidth}
  \vspace{-1em}
  \centering
  \small
  \setlength{\tabcolsep}{4pt}
  \renewcommand{\arraystretch}{1.08}

  \begin{tabularx}{\linewidth}{
    @{}
    >{\raggedright\arraybackslash}p{0.34\linewidth}
    >{\raggedright\arraybackslash}X
    @{}
  }
    \toprule
    \rowcolor{morandiheader}
    \textbf{Family} &
    \textbf{Model} \\
    \midrule

    \textbf{Image encoder}
      & DINOv2-Large \\

    \cmidrule(lr){1-2}

    \multirow{2}{*}{\textbf{Video encoder}}
      & V-JEPA2 ViT-L \\
      & VideoMAE-Large \\

    \cmidrule(lr){1-2}

    \textbf{3D encoder}
      & Fast3R ViT-L \\

    \cmidrule(lr){1-2}

    \multirow{4}{*}{\textbf{Video generator}}
      & Wan2.1-T2V-1.3B \\
      & CogVideoX-5B-I2V \\
      & MiniMax-H3 \\
      & LingBot-video MoE-30B-A3B \\

    \cmidrule(lr){1-2}

    \textbf{World model}
      & AetherV1 \\

    \cmidrule(lr){1-2}

    \textbf{Multimodal}
      & Qwen2.5-VL-7B-Instruct \\

    \bottomrule
  \end{tabularx}

  \caption{\textbf{Ten representations evaluated.}}
  \label{tab:model-overview}
  \vspace{-1.5em}
\end{wraptable}
We compare \textbf{ten frozen representations} spanning image encoders,
video encoders, 3D models, multimodal models, video generators, and world
models (\cref{tab:model-overview}); all backbone parameters remain frozen.
The representations are DINOv2 \citep{oquab2024dinov2}, V-JEPA2
\citep{assran2025vjepa2}, VideoMAE \citep{tong2022videomae}, Fast3R
\citep{yang2025fast3r}, Wan \citep{wanteam2025wan}, CogVideoX
\citep{yang2024cogvideox}, MiniMax-H3 \citep{minimax2026h3}, LingBot-video
\citep{ma2026lingbotvideo}, Aether \citep{zhu2025aether}, and Qwen2.5-VL
\citep{bai2025qwen25vl}.
Layer indices are model-specific and are not absolute measures of depth
across architectures. The designated representative layers are listed in
\cref{tab:extraction-details}.

Within a model and task, the readout architecture and objective are shared
across layers and prefix lengths. The main comparison averages over the
evaluated prefixes at the designated layer; layer and prefix sweeps are
analyzed separately rather than contributing their post-hoc maxima as
primary scores.

\paragraph{Readout metrics.}
Instance identity uses a cross-entropy loss on per-pixel embeddings, with
prototypes computed from ground-truth instance groups within the evaluated
sample; it measures identity separation, not mask quality or segmentation
IoU. Direct prediction and path integration retrieve future features within
each model's own target space, so their scores do not lie on a common scale
of physical prediction error. Global R@1 is the fraction of predictions
whose most similar candidate is the correct future feature. Camera-pose
AUC@30 is the area under the pose-accuracy curve up to a 30-degree error
threshold. Probes in the main comparison use training seed 42. Intervals
are 95\% scene-clustered bootstrap intervals over the 69 validation scenes
(1{,}000 resamples shared by all cells, so model, prefix, and coordinate
differences are paired); they reflect evaluation sampling, not
training-seed variability (\cref{sec:bootstrap-intervals}).

\paragraph{Evaluation scope.}
The main table reports final-checkpoint evaluations on 276 validation
windows per prefix. The MiniMax-H3 prediction probes are evaluated on 171
of these windows and are marked accordingly. Direct prediction for the
other nine models pools three future targets per window into a gallery of
828 candidates, whereas MiniMax-H3 direct prediction uses 171
next-observation candidates. Path integration pools the same three
waypoints, giving 828 candidates for the other models and 513 for
MiniMax-H3. Object persistence scores only windows with an eligible,
previously visible target: 80 windows at prefix 8 and 171 at each of
prefixes 12, 16, and 24. \Cref{sec:common-validation-subset} reports
results on the 171 shared validation windows together with a matched
next-observation evaluation. The matched input controls, the
native-prediction readout, the motion-composition study, and the action
experiments follow their own protocols, described below.

\section{Experiments}
\label{sec:experiments}
We first compare spatial readouts across the ten representations and
examine their dependence on the extraction setting
(\cref{sec:main-benchmark,sec:layer-prefix-analysis}). We then test
whether the probes use history and motion (\cref{sec:probe-validity}),
whether spatial readout survives native prediction
(\cref{sec:predictive-spatial-interface}), and whether goal information
has functional utility for action (\cref{sec:action-output}).

\subsection{Spatial capabilities differ across representations}
\label{sec:main-benchmark}
The main results are collected in \cref{tab:main-results}. For model $m$
and task $t$, the reported score averages over the evaluated prefixes
$\mathcal{P}_{m,t}$ at the designated layer $\ell_m$:
\begin{equation}
  \bar{s}_{m,t}=\frac{1}{|\mathcal{P}_{m,t}|}
  \sum_{p\in\mathcal{P}_{m,t}}s_{m,t}(\ell_m,p).
  \label{eq:main-prefix-average}
\end{equation}
No best-layer or best-prefix selection enters this aggregation.
\begin{table}[tbp]
  \centering
  \caption{\textbf{Main results for all ten representations.}
  Prefix aggregation follows \cref{eq:main-prefix-average} at each
  designated layer. The first header row names the diagnostic task and the
  second row its metric; metric geometry contributes two metrics.
  \textbf{Bold}, colored values mark the best result in each column and
  every result whose paired 95\% scene-bootstrap interval of the difference
  from it contains zero; marked cells are excluded. $\dagger$: MiniMax-H3
  prediction is evaluated on 171 of the 276 validation windows, and its
  direct predictor uses the next-observation target only.
  \Cref{tab:main-results-ci} gives the intervals of every cell and
  \cref{tab:main-results-common171} the shared-window comparison.}
  \label{tab:main-results}

  \setlength{\tabcolsep}{3pt}
  \footnotesize

  \begin{tabular}{@{}lrrrrrrr@{}}
    \toprule
    \cellcolor{headergray} &
    \multicolumn{2}{c}{\cellcolor{structblue}\taskgeometry} &
    \cellcolor{structblue}\makecell{Instance\\identity} &
    \cellcolor{structblue}\makecell{View\\consistency} &
    \cellcolor{memorygreen}\makecell{Object\\persistence} &
    \cellcolor{predictorange}\makecell{Direct\\prediction} &
    \cellcolor{predictorange}\makecell{Path\\integration} \\
    \cmidrule(lr){2-3}\cmidrule(lr){4-4}\cmidrule(lr){5-5}
    \cmidrule(lr){6-6}\cmidrule(lr){7-7}\cmidrule(lr){8-8}
    \cellcolor{headergray}Model (layer) &
    \cellcolor{structblue}\makecell{Depth\\loss $\downarrow$} &
    \cellcolor{structblue}\makecell{AUC@30\\$\uparrow$} &
    \cellcolor{structblue}\makecell{Instance\\loss $\downarrow$} &
    \cellcolor{structblue}\makecell{Bal.\\acc.\ $\uparrow$} &
    \cellcolor{memorygreen}\makecell{Ang.\ err.\\($^{\circ}$) $\downarrow$} &
    \cellcolor{predictorange}\makecell{Global\\R@1 $\uparrow$} &
    \cellcolor{predictorange}\makecell{Global\\R@1 $\uparrow$} \\
    \midrule
    \model{DINOv2-L} (0)
      & \textcolor{structink}{\textbf{0.1422}}
      & 0.0412
      & \textcolor{structink}{\textbf{1.3828}}
      & 0.9557
      & 65.38
      & \textcolor{predictink}{\textbf{0.2373}}
      & 0.0900 \\

    \model{Fast3R} (24)
      & \textcolor{structink}{\textbf{0.1386}}
      & \textcolor{structink}{\textbf{0.3870}}
      & 1.6072
      & \textcolor{structink}{\textbf{0.9808}}
      & \textcolor{memoryink}{\textbf{32.79}}
      & 0.0133
      & 0.0190 \\

    \model{VideoMAE-L} (23)
      & 0.2968
      & 0.2026
      & 4.1571
      & 0.9118
      & 36.45
      & 0.0531
      & 0.0685 \\

    \model{V-JEPA2-L} (23)
      & 0.2708
      & 0.1854
      & 2.9211
      & 0.9256
      & 45.31
      & 0.0625
      & \textcolor{predictink}{\textbf{0.1588}} \\

    \model{Qwen2.5-VL} ($-1$)
      & 0.2244
      & 0.0269
      & 2.9476
      & 0.9385
      & 68.59
      & 0.1712
      & 0.0903 \\

    \model{Wan} (20)
      & 0.3609
      & 0.0650
      & 4.9301
      & 0.8215
      & 62.38
      & 0.0260
      & 0.0405 \\

    \model{CogVideoX} (20)
      & 0.4340
      & 0.0276
      & 6.6992
      & 0.8001
      & 70.62
      & 0.0088
      & 0.0160 \\

    \model{Aether} (1)
      & 0.5270
      & 0.0185
      & 7.1847
      & 0.6928
      & 61.18
      & 0.0024
      & 0.0042 \\

    \model{MiniMax-H3} (30)
      & 0.4252
      & 0.0477
      & 6.2265
      & 0.8255
      & 55.97
      & 0.0249$^{\dagger}$
      & 0.0117$^{\dagger}$ \\

    \model{LingBot-video} (30)
      & 0.4314
      & 0.0232
      & 5.5875
      & 0.7858
      & 70.30
      & 0.0302
      & 0.0429 \\
    \bottomrule
  \end{tabular}
\end{table}

\paragraph{Spatial structure exposes distinct task strengths.}
Fast3R and DINOv2 illustrate that metric geometry, view consistency, and
instance identity are not interchangeable readouts. Fast3R has the highest
camera-pose AUC@30 and view accuracy in the main table, but its advantage
over DINOv2 is far larger for camera pose (0.3870 versus 0.0412) than for
depth loss or view accuracy, and DINOv2 has the lower instance-identity
loss (1.3828 versus 1.6072). Strong view consistency or instance
separation therefore does not imply equally strong camera-pose readout.

\paragraph{Visible-scene accuracy does not determine object persistence.}
Fast3R achieves the lowest object-persistence error (32.79$^{\circ}$),
followed by VideoMAE (36.45$^{\circ}$). VideoMAE localizes far more
accurately than DINOv2 (65.38$^{\circ}$) and Qwen2.5-VL
(68.59$^{\circ}$), although both outperform it on depth, instance
identity, and view consistency. Object persistence thus reveals differences
that visible-scene readout alone would miss, without requiring a different
winner on every task.

\paragraph{Object persistence aligns with camera-pose readout.}
The exception among the L1 metrics is camera pose. The three most accurate
localizers (Fast3R, VideoMAE, V-JEPA2) are exactly the three
representations with the highest AUC@30, and across all ten representations
the rank correlation between persistence and camera-pose readout is
$\rho=0.76$, compared with $\rho\le0.43$ for depth, instance identity, and
view consistency (Spearman, direction-corrected). Depth, instance identity,
and view consistency rank the models almost identically ($\rho\ge0.91$
pairwise), whereas camera pose does not ($\rho\le0.62$). Localizing an
object that has left the view requires relating the current camera to
earlier viewpoints, so a representation that exposes camera pose has the
ingredient that persistence demands. The layer sweeps corroborate this
within models: in all four multi-layer backbones (VideoMAE, V-JEPA2, Wan,
and CogVideoX), the block at which persistence peaks is the block at which
metric geometry peaks (\cref{fig:peak-layer-map}).

\paragraph{Persistence and prediction yield different model orderings.}
DINOv2 leads direct prediction (R@1 0.2373), whereas V-JEPA2 leads path
integration (0.1588). Qwen2.5-VL ranks second in direct prediction and
ties DINOv2 for second in path integration (0.0903 versus 0.0900) despite
a higher object-persistence error than VideoMAE and V-JEPA2, and
DINOv2 and V-JEPA2 reverse order between direct prediction (0.2373 versus
0.0625) and path integration (0.0900 versus 0.1588). Over the nine
representations evaluated under the common prediction protocol, the two
prediction tasks rank models similarly ($\rho=0.93$), but neither ordering
is related to the persistence ordering ($|\rho|<0.1$). Localization,
direct prediction, and sequential integration therefore warrant separate
evaluation. MiniMax-H3 is left out of the prediction rankings because its
probes use a different window set and, for direct prediction, a different
target protocol. Against matched action controls
(\cref{tab:c1-action-controls-ci}), the direct predictor's gain over both
the no-action and the shuffled-action input is above zero for eight
models, though small for CogVideoX, Fast3R, and Wan; MiniMax-H3 exceeds
only the no-action control, and Aether's direct prediction is not
distinguishable from chance or from either control.

\paragraph{Representation families do not determine capability profiles.}
VideoMAE and V-JEPA2 have close view-consistency accuracy (0.9118 versus
0.9256) but different strengths: VideoMAE is better on camera pose and
object persistence, whereas V-JEPA2 is better on depth and path
integration; their direct-prediction scores lie within each other's
interval. Among the generative and world-model representations, Aether
has a lower object-persistence error than CogVideoX and matches Wan
(61.18$^{\circ}$ versus 62.38$^{\circ}$) despite weaker results on the
other six metrics, and MiniMax-H3 localizes best within this group
(55.97$^{\circ}$). These comparisons describe the selected representations
and extraction points; they do not isolate the effect of the training
objective, since architecture and training data differ as well.

\begin{keyfinding}{warmgray}{No task-independent model ranking}
The observed advantages depend on the spatial function being tested.
Strong visible-scene readout does not guarantee accurate object persistence
or strong direct prediction and path integration, and the persistence and
prediction orderings are unrelated. A single diagnostic therefore cannot
summarize a representation's spatial capability profile.
\end{keyfinding}

\subsection{Spatial accessibility depends on layer and streaming prefix}
\label{sec:layer-prefix-analysis}

\begin{figure}[tbp]
  \centering
  \begin{subfigure}[t]{0.46\linewidth}
    \centering
    \includegraphics[width=\linewidth]{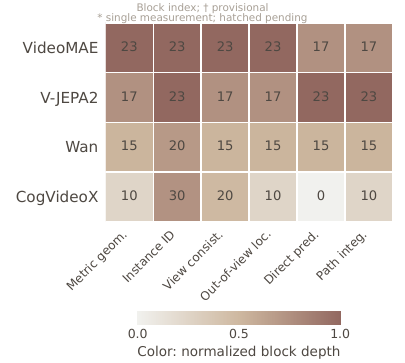}
    \caption{Best observed layers.}
    \label{fig:peak-layer-map}
  \end{subfigure}\hspace{0.03\linewidth}
  \begin{subfigure}[t]{0.46\linewidth}
    \centering
    \includegraphics[width=\linewidth]{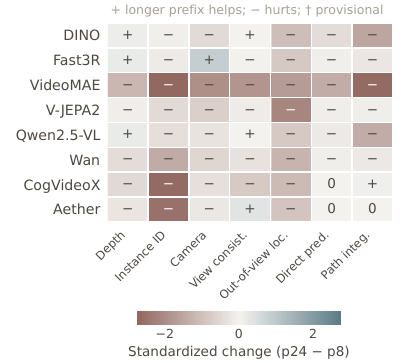}
    \caption{Effects of a longer prefix.}
    \label{fig:prefix-integration}
  \end{subfigure}
  \caption{\textbf{Spatial readout depends on both layer and streaming prefix.}
  (a) Four multi-layer backbones: cells give the best measured block,
  colored by normalized depth. Metric geometry averages the percentile
  ranks of depth and camera-pose performance.
  (b) Paired comparisons for eight models at their designated layers:
  direction-corrected, standardized p24$-$p8 differences following
  \cref{eq:prefix-effect}; $+$ favors the longer prefix and $-$ the
  shorter.
  Direct prediction pools horizons 1, 2, and 4 for these models. Full
  layer curves and joint grids are given in \cref{sec:additional-figures}.}
  \label{fig:layer-history-summary}
\end{figure}

\paragraph{Different readouts favor different layers.}
\Cref{fig:peak-layer-map} summarizes the best measured layer for the
backbones with multi-layer sweeps. V-JEPA2 peaks at block 23 for direct
prediction, path integration, and instance identity, but at block 17 for
object persistence; VideoMAE instead peaks at block 23 for localization
and instance identity and at block 17 for both prediction tasks. The two
video generators peak earlier in the network: Wan's best block is 15 for
five of the six readouts, although its designated main-table layer is
block 20, and CogVideoX peaks at block 10 or earlier for metric geometry,
object persistence, and both prediction tasks. The layer sweeps in
\cref{fig:layer-emergence} show the same contrast in profile shape: the
two video encoders improve monotonically with depth on the L1 readouts,
whereas the generators' metric-geometry and view-consistency curves peak
at intermediate blocks and decline toward the output. A
representative layer selected for one task therefore need not maximize
another task's readout, and cross-model orderings can change with the
setting: at block 17 and prefix 8, V-JEPA2 reaches a localization error of
28.8$^{\circ}$, below VideoMAE's best cell (29.4$^{\circ}$;
\cref{fig:layer-prefix-persistence}), whereas the designated layers favor
VideoMAE by about 9$^{\circ}$.

\paragraph{Longer prefixes do not uniformly improve spatial readout.}
We compare p24 with p8 using direction-corrected standardized differences:
\begin{equation}
  \Delta_{m,t}=
  \frac{d_t\,[s_{m,t}(\ell_m,24)-s_{m,t}(\ell_m,8)]}{\sigma_t},
  \label{eq:prefix-effect}
\end{equation}
where $d_t=+1$ for higher-is-better metrics and $-1$ otherwise, and
$\sigma_t$ is the standard deviation of the metric across all
model--layer--prefix cells. Among the 56 model--metric endpoint pairs for
the eight models in \cref{fig:prefix-integration}, 45 favor p8, eight
favor p24, and three tie; 33 and five of these, respectively, have paired
95\% intervals that exclude zero (\cref{tab:prefix-effects-ci}). All eight
object-persistence pairs worsen with the longer prefix, each beyond its
interval, and for path integration six models worsen, three of them
(DINOv2, Qwen2.5-VL, and VideoMAE) beyond their intervals, CogVideoX
improves within its interval, and Aether ties.
For example, VideoMAE's localization error rises from 29.55$^{\circ}$ to
54.80$^{\circ}$, and DINOv2's path-integration R@1 falls from 0.1256 to
0.0713. Every pair that favors p24 belongs to a visible-scene readout
(depth, camera pose, or view consistency), with the single exception of
CogVideoX path integration; Fast3R, for instance, improves on depth and
camera pose. The shape of the degradation also differs between models:
VideoMAE's block-23 readouts are stable up to p16 and collapse at p24
(instance loss 2.47 to 9.98, localization 32.1$^{\circ}$ to
54.8$^{\circ}$), whereas V-JEPA2 degrades more gradually across prefixes
(\cref{fig:layer-prefix-persistence}). Because a longer prefix also moves
the endpoint (\cref{sec:streaming-evaluation}), these differences do not
isolate history alone.

\begin{keyfinding}{structblue}{Spatial accessibility is setting-dependent}
Different spatial readouts favor different layers, and longer visual
prefixes do not uniformly improve them. The extraction setting is part of
the evaluation, not an interchangeable choice of feature interface.
\end{keyfinding}

\subsection{What do the probes actually use?}
\label{sec:probe-validity}

\begin{table}[tbp]
  \centering
  \caption{\textbf{Motion and history controls.}
  Motion gain is the correct-minus-wrong-motion R@1 in the
  matched-candidate protocol; history benefit is the reduction in
  localization error from full history relative to the current frame
  alone. Brackets give paired 95\% scene-bootstrap intervals.}
  \label{tab:probe-validity-controls}
  \small
  \setlength{\tabcolsep}{6pt}
  \renewcommand{\arraystretch}{1.10}
  \begin{tabular}{@{}lrr@{}}
    \toprule
    \rowcolor{morandiheader}
    \textbf{Model} &
    \makecell[r]{\textbf{Motion gain}\\\textbf{(pp)}} &
    \makecell[r]{\textbf{History benefit}\\\textbf{($^\circ$)}} \\
    \midrule
    V-JEPA2
      & \makecell[tr]{18.84\\ \scriptsize [13.04, 25.36]}
      & \makecell[tr]{10.26\\ \scriptsize [3.51, 16.80]} \\
    \addlinespace[2pt]
    VideoMAE
      & \makecell[tr]{10.51\\ \scriptsize [6.16, 15.22]}
      & \makecell[tr]{2.08\\ \scriptsize [\negnum{6.10}, 9.97]} \\
    \addlinespace[2pt]
    Wan
      & \makecell[tr]{2.90\\ \scriptsize [\negnum{1.45}, 7.25]}
      & \makecell[tr]{10.01\\ \scriptsize [3.42, 16.69]} \\
    \addlinespace[2pt]
    CogVideoX
      & \makecell[tr]{4.35\\ \scriptsize [0.35, 7.97]}
      & \makecell[tr]{\negnum{3.41}\\ \scriptsize [\negnum{9.39}, 2.39]} \\
    \bottomrule
  \end{tabular}
\end{table}

We complement the main benchmark with separate baseline comparisons and
input controls that examine the contributions of motion and visual
history. These matched-control experiments have their own evaluation
windows and candidate pools, so their retrieval percentages are not the
mixed-horizon global R@1 values of \cref{tab:main-results}. In every
control, the trained probe and the prediction targets remain fixed, which
allows a direct comparison between full and modified conditioning inputs.

\paragraph{One-step prediction controls.}
In a one-step protocol (\cref{sec:protocol-details}), we first compare
learned prediction with copying the last observed feature as a temporal
baseline. In global future-feature retrieval, Wan achieves 15.22\% with
prediction and 11.96\% with copying, whereas V-JEPA2 achieves 13.04\% and
19.93\%, respectively: for V-JEPA2, copying exceeds learned prediction. We
then evaluate motion conditioning with eight pose-displacement-matched
candidates. Correct motion improves retrieval over matched wrong motion
for V-JEPA2, VideoMAE, and CogVideoX, with gains of 18.84, 10.51, and 4.35
percentage points; Wan's gain of 2.90 points has an interval spanning zero
(\cref{tab:probe-validity-controls}). The copying comparison provides a
reference for global retrieval, whereas the matched-candidate comparison
quantifies the benefit of the correct motion. A complementary action-set
evaluation of the main direct predictors is reported in
\cref{sec:direct-action-sensitivity}.

\paragraph{Object persistence.}
We compare localization from the full visual history with localization
from the current frame alone, retaining the same historical object query
in both conditions. Full history reduces the angular error of V-JEPA2 and
Wan by 10.26$^\circ$ and 10.01$^\circ$, respectively, with both intervals
above zero. The corresponding effects for VideoMAE and CogVideoX are
smaller and have intervals spanning zero
(\cref{tab:probe-validity-controls}); removing the target-visible frames,
zeroing all visual tokens, and shuffling or reversing the history are
reported in \cref{tab:history-interventions}. Notably, VideoMAE, the most accurate
of these four localizers in \cref{tab:main-results}, shows no measurable
benefit from history beyond the current frame, so a strong localization
score does not by itself establish that the probe reads out accumulated
history.

\paragraph{Motion-composition consistency.}
We additionally examine how an external recurrent updater trained on
one-step transitions generalizes to the composition of two motions. For
Wan and V-JEPA2, updating improves endpoint prediction over copying.
Correct-order effects remain inconclusive for both the feature-based
updaters and a learned geometric control, whereas an analytic SE(3)
reference preserves transformation order. The full comparison is given in
\cref{sec:se3-diagnostic-details}.

\begin{keyfinding}{memorygreen}{Benefits from motion and visual history}
Correct motion improves matched-candidate prediction for V-JEPA2,
VideoMAE, and CogVideoX, and visual history improves localization for
V-JEPA2 and Wan, but not measurably for VideoMAE. These controls
complement the main capability profiles with measured gains from
task-relevant inputs and show that decodability and input use can
dissociate.
\end{keyfinding}

\subsection{Prediction changes the spatial readout interface}
\label{sec:predictive-spatial-interface}
\label{sec:representation-specific}
\begin{figure}[tbp]
  \centering
  \begin{subfigure}[t]{0.46\linewidth}
    \centering
    \includegraphics[width=\linewidth]{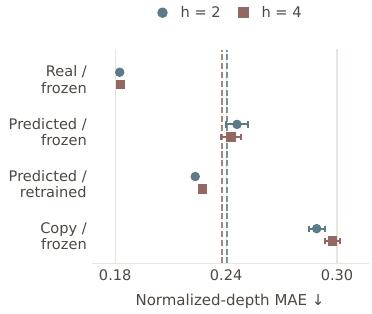}
    \caption{Readout through native prediction.}
    \label{fig:predictive-spatial-interface}
  \end{subfigure}\hspace{0.03\linewidth}
  \begin{subfigure}[t]{0.46\linewidth}
    \centering
    \includegraphics[width=\linewidth]{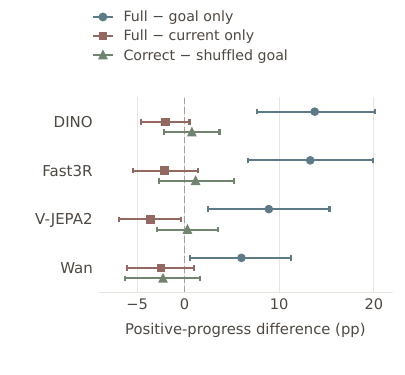}
    \caption{Goal-specific utility.}
    \label{fig:goal-conditioning}
  \end{subfigure}
  \caption{\textbf{Testing readable structure and goal-specific utility.}
  (a) V-JEPA2 normalized-depth MAE at $h=2$ and $h=4$ with the same
  frozen reader, except for the retrained control; dashed lines mark the
  training-layout priors, and error bars show the SD over three seeds.
  (b) Positive-progress differences in the continuous-output setting under
  input restriction (full minus goal-only, full minus current-only) and
  fixed-weight goal shuffling (correct minus shuffled goal); error bars
  show paired 95\% scene-bootstrap intervals.}
  \label{fig:prediction-action-controls}
\end{figure}

As a further L3 analysis, we ask whether the L1 metric geometry that is
accessible in real future features remains accessible after native
prediction. A probe trained on real V-JEPA2 future states is applied
unchanged to the model's native predictions
(\cref{fig:predictive-spatial-interface}). Supporting controls first
establish target-specific real-feature readout and motion-conditioned
utility for an external latent predictor; details are given in
\cref{sec:predictive-readout-controls}.

At horizon two, normalized-depth MAE rises from 0.182 on real features to
0.246 on native predictions. Retraining an equal-capacity reader on the
predictions improves the MAE to 0.223, better than the 0.241 layout prior
but worse than the real-feature readout; horizon four shows the same
pattern. Frozen transfer remains better than copied context but slightly
worse than the prior (\cref{tab:predictive-spatial-interface}). The
results distinguish recoverable spatial structure from preservation of the
original readout interface.

\begin{keyfinding}{predictorange}{Recoverable structure is not unchanged readability}
Natively predicted features retain some recoverable spatial structure, but
a reader trained on real features does not keep its accuracy when applied
to them. Prediction supports neither lossless spatial propagation nor the
claim that all spatial information has disappeared.
\end{keyfinding}

\subsection{Goal-conditioned action prediction}
\label{sec:action-output}

We test actionability through \textbf{goal-conditioned action prediction}
in two complementary settings. The candidate-selection setting chooses
between supplied actions, whereas the continuous-output setting predicts
an action directly. Both use p8 features and an independently extracted
goal feature.

\begin{table}[tbp]
  \centering
  \caption{\textbf{Goal-conditioned action prediction: candidate-selection
  setting.} Accuracy (\%), mean$\pm$SD over three training seeds; chance
  is 50\%. Shuffled goal evaluates the trained direct scorer with mismatched
  goals at fixed weights. The endpoint reference observes realized outcomes
  and is diagnostic only.}
  \label{tab:action-output}
  \small
  \setlength{\tabcolsep}{5pt}
  \renewcommand{\arraystretch}{1.15}
  \begin{tabularx}{0.9\linewidth}{
    l
    >{\raggedleft\arraybackslash}X
    >{\raggedleft\arraybackslash}X
    >{\raggedleft\arraybackslash}X
    >{\raggedleft\arraybackslash}X
    >{\raggedleft\arraybackslash}X
  }
    \toprule
    \rowcolor{morandiheader}
    \textbf{Representation}
      & \makecell[r]{\textbf{Action}\\\textbf{only}}
      & \textbf{Direct}
      & \makecell[r]{\textbf{Shuffled}\\\textbf{goal}}
      & \makecell[r]{\textbf{Predict-}\\\textbf{then-score}}
      & \makecell[r]{\textbf{Endpoint}\\\textbf{ref.}} \\
    \midrule

    DINOv2-L
      & $67.8{\pm}0.4$
      & $76.2{\pm}0.8$
      & $55.5{\pm}0.8$
      & $76.2{\pm}1.2$
      & $85.5$ \\

    Fast3R
      & $67.8{\pm}0.4$
      & $79.9{\pm}1.3$
      & $62.4{\pm}0.5$
      & $78.1{\pm}0.3$
      & $78.0$ \\

    V-JEPA2
      & $67.8{\pm}0.4$
      & $78.1{\pm}0.7$
      & $62.0{\pm}1.1$
      & $76.7{\pm}0.8$
      & $78.0$ \\

    Wan
      & $67.8{\pm}0.4$
      & $75.9{\pm}0.9$
      & $63.7{\pm}0.5$
      & $63.6{\pm}1.1$
      & $68.3$ \\

    \bottomrule
  \end{tabularx}
\end{table}

\paragraph{Candidate-selection setting.}
A shallow probe chooses between two 9D camera motions according to their
progress toward the goal, where progress is the decrease in the pose
distance to the goal, defined as the translation distance plus 0.5 times
the rotation angle in radians. Direct selection reaches 75.9--79.9\%
accuracy, exceeding the
67.8\% action-only baseline, and shuffling the goals reduces accuracy by
12.2--20.7 percentage points (\cref{tab:action-output}). The readout
therefore uses goal information for candidate selection. Predicting each
candidate's outcome before scoring it offers no consistent advantage and
reduces Wan's accuracy to 63.6\%.

\paragraph{Continuous-output setting.}
Without candidates, a shallow probe predicts the first observed 9D motion
of a four-step trajectory from the current p8 feature and an isolated
horizon-four goal feature. Separate projections $c$ and $g$ are fused as
$[c,g,g-c,c\odot g]$ before prediction; the matched-capacity concatenation
control is specified in \cref{sec:action-protocol-details}. FULL achieves
positive progress on 50.2--59.8\% of trajectories
(\cref{tab:continuous-action}) and exceeds GOAL-ONLY by 6.0--13.8
percentage points, but it does not improve positive progress over
CURRENT-ONLY for any backbone (\cref{fig:goal-conditioning}).
Beating GOAL-ONLY establishes that the current state is useful; it does
not establish a benefit from adding the goal as an input. Fast3R, the
representation with the strongest camera-pose and persistence readouts,
attains the highest positive-progress rate (59.8\%) and is the only
representation whose FULL$-$CURRENT-ONLY interval for goal reduction
includes zero.

\paragraph{Changing the goal changes actions, not reliably their utility.}
Holding the current state and the weights fixed while substituting a
difficulty-matched goal from a different scene changes the predicted
direction by 29.8--45.6$^{\circ}$. Yet the correct-minus-shuffled-goal
effects on positive progress ($-2.3$ to $+1.2$ percentage points) and on
goal reduction ($-0.006$ to $+0.001$) all have 95\% intervals containing
zero (\cref{fig:goal-conditioning}). Goal sensitivity therefore does not
translate into reliable goal-specific utility under this readout. The
relational head improves some metrics over concatenation but does not
close this gap. These are supervised offline diagnostics, not a
demonstration of a closed-loop policy.

\begin{keyfinding}{morandisand}{Goal sensitivity is not goal utility}
Changing the goal changes the predicted actions, but the correct goal does
not reliably improve progress. Success in the candidate-selection setting
therefore does not establish reliable goal-conditioned action prediction
in the continuous-output setting.
\end{keyfinding}

\subsection{How the four levels relate}
\label{sec:level-relations}

\paragraph{Persistence builds on camera-pose readout.}
As shown in \cref{sec:main-benchmark}, the three most accurate localizers
are the three representations with the highest camera-pose AUC@30, the
cross-model rank correlation between the two is $\rho=0.76$, and within
VideoMAE, V-JEPA2, and Wan the persistence peak coincides with the
metric-geometry peak across layers. No representation with weak
camera-pose readout localizes well.

\paragraph{Prediction builds on visible-scene readout, not on persistence.}
Direct prediction and path integration correlate with depth, instance
identity, and view consistency ($\rho=0.52$--$0.63$ over the nine
representations under the common prediction protocol) but not with camera
pose ($\rho\le0.25$) or persistence ($|\rho|<0.1$). The four best
predictors on both tasks (DINOv2, Qwen2.5-VL, V-JEPA2, VideoMAE) are, after
Fast3R, the four best representations on depth and view consistency.
Fast3R itself, the best on every L1 metric except instance identity, ranks
eighth on direct prediction and seventh on path integration: strong
visible-scene readout is necessary for feature prediction but not
sufficient.

\paragraph{Actionability follows the persistence chain.}
Among the four representations evaluated at L4, Fast3R leads both action
settings (79.9\% candidate-selection accuracy and 59.8\% positive
progress), V-JEPA2 is second in candidate selection (78.1\%) and tied with
DINOv2 in positive progress (53.8\%), and Wan is last in both (75.9\% and
50.2\%). This is the order of object persistence (32.79$^{\circ}$,
45.31$^{\circ}$, then 62.38$^{\circ}$ and 65.38$^{\circ}$) and of
camera-pose readout within the group, not the order of direct prediction,
where DINOv2 is first (0.2373) and Fast3R last (0.0133). Goal-conditioned
action therefore tracks the pose-and-persistence chain rather than the
prediction chain.

\begin{keyfinding}{headergray}{Two ingredient chains run through the levels}
Camera-pose readout supports object persistence, which in turn supports
goal-conditioned action; visible-scene readout supports feature
prediction. In each chain the lower level bounds the higher one without
guaranteeing it, which is why a single diagnostic cannot stand in for the
profile.
\end{keyfinding}

\section{Related Work}
\label{sec:related-work}

\paragraph{Probing spatial and 3D structure in frozen representations.}
Frozen-feature probing asks what information is already accessible in a
representation without updating the backbone. Self-supervised image
encoders such as DINO and DINOv2 revealed that semantic organization,
object boundaries, and strong local correspondences emerge without dense
labels \citep{caron2021dino,oquab2024dinov2}. More recent work extends
this methodology to video foundation models: VidFM3D evaluates whether
shallow readouts recover depth, camera pose, and 3D point structure from
frozen video-model features, and analyzes how these signals vary with
network depth and diffusion timestep \citep{huang2025vidfm3d}. Related
probes of image foundation models test depth, surface normals, and
multi-view correspondence \citep{elbanani2024probing}, scene-level tasks
such as 3D question answering, visual grounding, and registration
\citep{man2024lexicon3d}, and novel-view synthesis from features read out
as 3D Gaussians \citep{chen2025feat2gs}. Feed-forward 3D reconstruction
models, from DUSt3R to VGGT and Fast3R, provide a geometry-specialized
reference point because their training directly targets multi-view
reconstruction and camera estimation
\citep{wang2024dust3r,wang2025vggt,yang2025fast3r}; streaming variants
such as Spann3R and CUT3R carry an explicit spatial memory or a persistent
state that is updated with every new frame
\citep{wang2025spann3r,wang2025cut3r}. Our streaming prefixes ask a
related question of general-purpose representations without assuming such
a state. Our work
starts from the same evidence-first principle but changes the object of
evaluation: metric geometry is one functional diagnostic rather than a
definition of spatial state, and we also ask whether a representation
supports view consistency, object persistence, predictive spatial update,
and actionability. Two models with similar depth readout can therefore
receive very different spatial-state profiles, and strong future-feature
retrieval is not credited as predictive spatial update unless it passes
matched motion controls.

\paragraph{Internal representations of diffusion and video generation models.}
Large video generators such as Wan and CogVideoX are optimized for
denoising and visual synthesis rather than for downstream readout
\citep{wanteam2025wan,yang2024cogvideox}. Nevertheless, work on image
diffusion models has shown that semantic and geometric consistency can be
extracted from internal activations, and that useful information is
distributed across both network layers and diffusion timesteps
\citep{tang2023dift,luo2023diffusionhyperfeatures}. Recent probing of
video diffusion transformers similarly reports that physical and
3D-related signals are most accessible at intermediate blocks and
nontrivial noise levels \citep{huang2025vidfm3d,esmati2026invisible}.
This motivates our layer $\times$ generation-coordinate $\times$
streaming-prefix analysis: we compare selected extraction settings rather
than assuming that one hidden state characterizes an entire model. Such
profiles describe accessibility during computation, not when a capability
emerges during training.

\paragraph{Spatial representations in foundation models.}
Spatial ability is often evaluated through language. SpatialVLM and
SpatialRGPT improve quantitative and region-level spatial question
answering by training vision-language models on large synthetic spatial
data \citep{chen2024spatialvlm,cheng2024spatialrgpt}, and VSI-Bench asks
multimodal LLMs to remember and recall the layout of a space from video,
finding that generating cognitive maps improves distance estimates
\citep{yang2025thinkinginspace}. Cambrian-S argues that such training
alone does not yield robust implicit spatial representations and proposes
predictive sensing, in which latent next-frame prediction decides what to
remember \citep{yang2025cambrians}. A growing set of studies probes the
representations themselves. SeeSE3 finds subspaces of self-supervised
vision features that align with 3D Euclidean space, with camera motion
linearly accessible from feature displacements \citep{chen2026seese3};
SpaRRTa shows that visual foundation models differ markedly in how well
they encode the relative positions of objects \citep{kargin2026sparrta};
probes of V-JEPA~2 and VideoMAE-v2 decode camera motion best at 60--70\%
of network depth while intuitive physics stays near chance
\citep{musa2026whatwherehow}, and in video encoders, motion direction
becomes decodable only within an intermediate-depth zone
\citep{joseph2026interpreting}. For multimodal LLMs, the spatial benefit of
a visual representation tracks its 3D structure, measured by the
cross-view consistency of latent features or the accuracy of explicit
geometry \citep{huang2026goodvisualrep}. Our study complements these probes by reading
several functional levels of spatial state, including object persistence,
predictive update, and goal use, from the same ten representations under
matched controls.

\paragraph{World models and embodied representations.}
Video world models aim to support interaction and planning. Genie learns
an action-controllable interactive environment from unlabeled Internet
videos \citep{bruce2024genie}; Navigation World Models predict future
observations from past observations and navigation actions and use them to
plan \citep{bar2025nwm}; DINO-WM and V-JEPA~2 plan with predictors in a
pre-trained or self-supervised feature space
\citep{zhou2025dinowm,assran2025vjepa2}; and Aether jointly learns 4D
reconstruction, action-conditioned video prediction, and goal-conditioned
visual planning \citep{zhu2025aether}. Whether video generators capture
physical regularities remains contested: diffusion models trained on
simulated motion generalize within the training distribution but fail
outside it \citep{kang2025howfar}. For embodied learning, SPA shows that
adding 3D spatial awareness to a vision encoder improves downstream policy
learning across many simulated tasks \citep{zhu2025spa}. Our L3 and L4
probes ask a narrower question of frozen features: whether ego-motion
updates and goal-conditioned actions can be read out from the represented
state.

\paragraph{Object-centric representations and persistence.}
Object-centric learning explicitly decomposes scenes into entity-level
latent variables. Slot Attention learns exchangeable slots that bind to
objects, and sequential extensions such as SAVi propagate and correct
object slots through video \citep{locatello2020slotattention,kipf2022savi}.
These approaches build an object-centric format into the architecture,
whereas the foundation models we evaluate generally use distributed patch
or video tokens, and we do not assume that their internal state is
slot-based. Our question is operational: can an object's bearing be
recovered after it is no longer visible? This differs from
violation-of-expectation benchmarks such as IntPhys~2, which assess the
discrimination of possible from impossible events involving permanence,
continuity, solidity, or immutability \citep{bordes2025intphys2}; V-JEPA
surprise scores perform well on several such conditions
\citep{garrido2025intuitivephysics}, but plausibility discrimination does
not measure object persistence directly. In generated video, current world
models usually resume a target that comes back into view in the state it
was left in, rather than letting its event progress while it was unseen
\citep{lu2026persistent}. We therefore predict an explicit
bearing in the current camera frame and use history interventions to test
which visual evidence supports that readout.

\section{Discussion and Conclusion}
\label{sec:conclusion}

We characterized spatial state representations through four functional
levels: spatial structure, object persistence, predictive spatial update,
and actionability. Across ten frozen representations, no model leads on
every task, and the readouts vary with the extraction layer and the
streaming prefix. The levels are linked by two ingredient chains:
camera-pose readout supports object persistence, which in turn supports
goal-conditioned action, whereas feature prediction builds on
visible-scene readout and is unrelated to persistence. Matched controls
show that decodability, input use, and functional utility dissociate: a
probe can localize accurately without using history, a natively predicted
feature retains recoverable structure but not the original readout
interface, and a goal-sensitive action readout need not benefit from the
correct goal. Together, these results support multidimensional evaluation
of spatial state beyond a single task score. Our conclusions concern
frozen feature interfaces, trained probes, and offline indoor
trajectories: independently encoded causal prefixes characterize what a
representation of the observed history affords rather than an incremental
state update, and retrieval scores in different target feature spaces do
not share a physical error scale.

\paragraph{Acknowledgments.}
We thank Saining Xie for valuable suggestions on this paper. We also
thank Weiyang Jin, Yuwei Niu, Feiyang Liu, Jiacheng Dong, Yuxuan Cheng,
and Yuhao Zhang for helpful discussions.

\clearpage

\bibliographystyle{plainnat}
\bibliography{iclr2027_conference,refs_spatial_20261029}

\clearpage

\appendix
\section{Appendix}
\suppressfloats[t]
\subsection{Additional layer, history, and intervention analyses}
\label{sec:additional-figures}

The main text keeps compact summaries. Here we provide the full measured
layer profiles, the joint instance-identity and localization grids, and
the intervention schematic; the generation-coordinate grid follows in
\cref{sec:generation-coordinate-details}. Layer scans complement the
representative-layer benchmark: they describe extraction-depth profiles,
not training-time emergence.

\begin{figure}[tbp]
  \centering
  \includegraphics[width=0.98\textwidth]
    {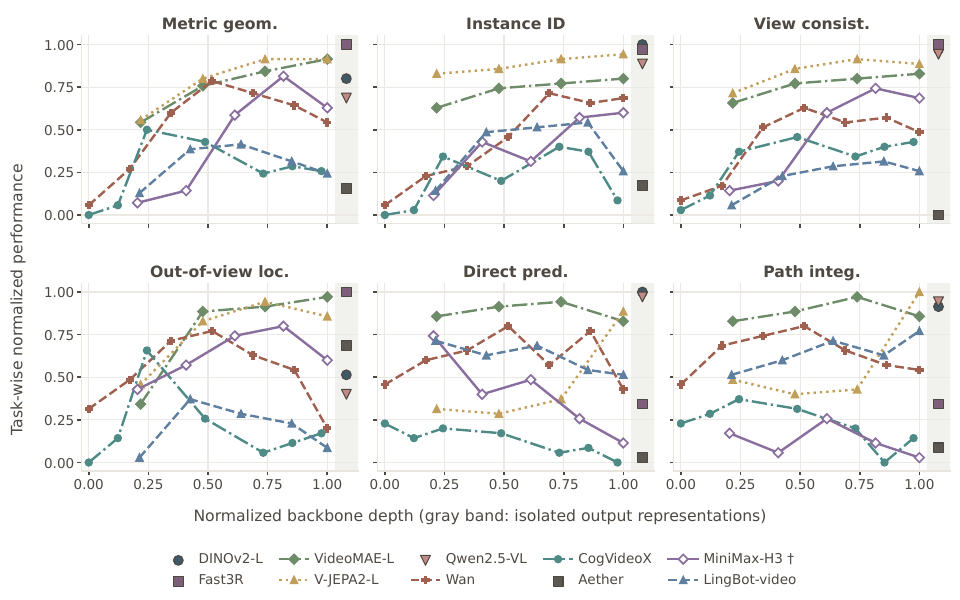}
  \caption{\textbf{Capability profiles across measured layers.}
  Prefix means are direction-corrected and converted to percentile ranks
  across the available model--layer cells within each task; metric
  geometry averages the depth and camera-pose ranks. The horizontal axis
  shows normalized backbone depth, and isolated output representations
  occupy a separate band on the right. Hollow markers denote the
  MiniMax-H3 points, which are evaluated on the 171-window subset (with
  $h=1$ only for direct prediction).}
  \label{fig:layer-emergence}
\end{figure}

\begin{figure}[tbp]
  \centering
  \includegraphics[width=0.98\textwidth]
    {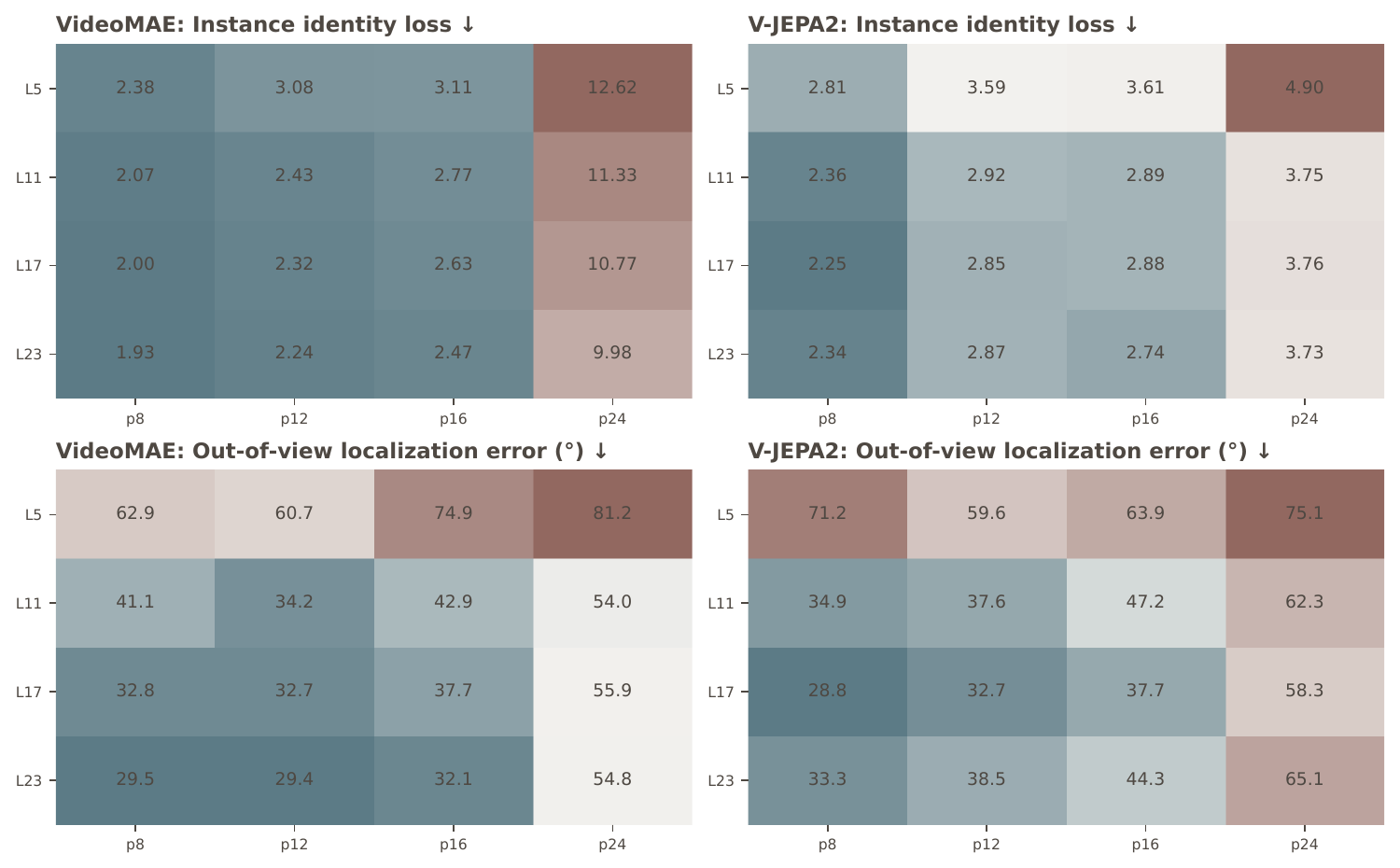}
  \caption{\textbf{Layer--prefix structure of instance identity and object persistence.}
  The four panels show instance identity (top) and object persistence
  (bottom) for VideoMAE (left) and V-JEPA2 (right) across extraction
  layers and prefix lengths. Colors are normalized within each panel and
  therefore describe within-model structure rather than absolute
  cross-model accuracy; raw values accompany the heatmaps. No probe was
  trained on block 0 of either model, so it is not shown.}
  \label{fig:layer-prefix-persistence}
\end{figure}

\begin{figure}[tbp]
  \centering
  \includegraphics[width=\linewidth]{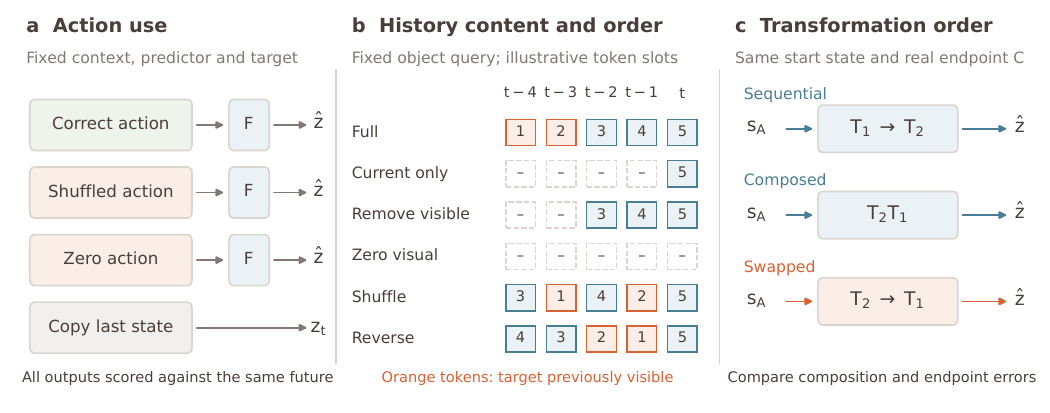}
  \caption{\textbf{Interventions separate information use from decodability.}
  (a) Change the action supplied to a fixed predictor $F$, or copy the last
  state. (b) Remove or reorder visual feature tokens while retaining the
  object query; five illustrative slots are shown, with temporal
  identifiers and the final frame fixed during order tests. (c) Compare
  sequential, composed, and swapped updates against the same real
  endpoint.}
  \label{fig:probe-interventions}
\end{figure}

\begin{table}[tbp]
  \centering
  \caption{\textbf{History interventions for object persistence.}
  Increase in localization error ($^\circ$) relative to the full history at
  p12 (165 samples from 65 scenes); brackets give paired 95\%
  scene-bootstrap intervals, and positive values mean that the intervention
  hurts. Every condition keeps the historical object query
  (\cref{fig:probe-interventions}b). The current-frame-only column repeats
  \cref{tab:probe-validity-controls}.}
  \label{tab:history-interventions}
  \footnotesize
  \setlength{\tabcolsep}{3pt}
  \begin{tabular}{@{}lrrrrr@{}}
    \toprule
    \rowcolor{morandiheader}
    \textbf{Model} & \makecell[r]{Current\\frame only} & \makecell[r]{Remove\\target-visible} & \makecell[r]{No visual\\tokens} & \makecell[r]{Shuffle\\order} & \makecell[r]{Reverse\\order} \\
    \midrule
    V-JEPA2
      & \makecell[tr]{10.26\\ \scriptsize [3.51, 16.80]}
      & \makecell[tr]{9.22\\ \scriptsize [3.95, 14.59]}
      & \makecell[tr]{9.66\\ \scriptsize [2.74, 16.85]}
      & \makecell[tr]{0.04\\ \scriptsize [\negnum{2.54}, 2.98]}
      & \makecell[tr]{2.15\\ \scriptsize [0.22, 4.56]} \\
    \addlinespace[2pt]
    VideoMAE
      & \makecell[tr]{2.08\\ \scriptsize [\negnum{6.10}, 9.97]}
      & \makecell[tr]{5.83\\ \scriptsize [\negnum{2.09}, 13.57]}
      & \makecell[tr]{6.06\\ \scriptsize [\negnum{2.36}, 14.57]}
      & \makecell[tr]{0.02\\ \scriptsize [\negnum{1.95}, 1.87]}
      & \makecell[tr]{1.09\\ \scriptsize [\negnum{1.25}, 3.07]} \\
    \addlinespace[2pt]
    Wan
      & \makecell[tr]{10.01\\ \scriptsize [3.42, 16.69]}
      & \makecell[tr]{6.59\\ \scriptsize [0.58, 12.91]}
      & \makecell[tr]{13.55\\ \scriptsize [7.31, 19.61]}
      & \makecell[tr]{0.95\\ \scriptsize [\negnum{1.85}, 3.94]}
      & \makecell[tr]{1.85\\ \scriptsize [\negnum{1.00}, 4.60]} \\
    \addlinespace[2pt]
    CogVideoX
      & \makecell[tr]{\negnum{3.41}\\ \scriptsize [\negnum{9.39}, 2.39]}
      & \makecell[tr]{0.54\\ \scriptsize [\negnum{4.33}, 5.53]}
      & \makecell[tr]{\negnum{0.83}\\ \scriptsize [\negnum{6.93}, 5.05]}
      & \makecell[tr]{0.00\\ \scriptsize [0.00, 0.00]}
      & \makecell[tr]{\negnum{0.13}\\ \scriptsize [\negnum{0.32}, 0.00]} \\
    \bottomrule
  \end{tabular}
\end{table}

\paragraph{History interventions.}
\Cref{tab:history-interventions} reports the interventions of
\cref{fig:probe-interventions}b together with the current-frame-only
condition. Removing the frames in which the target is visible, or zeroing
all visual tokens while keeping the object query, increases the error of
V-JEPA2 and Wan with intervals above zero, whereas VideoMAE and CogVideoX
show no reliable change. Shuffling the frame contents changes the error by
at most 0.95$^\circ$, with every interval containing zero, and leaves
CogVideoX's predictions unchanged. Reversing the history increases
V-JEPA2's error by 2.15$^\circ$ [0.22, 4.56] and changes the other models by
less than 2$^\circ$, with intervals containing zero. Because shuffling and
reversal keep the final frame and the temporal identifiers
(\cref{sec:protocol-details}), these interventions test the use of
historical content rather than ordered state updating.

\subsection{Comparison on shared validation windows}
\label{sec:common-validation-subset}
\Cref{tab:main-results-common171} restricts each model to the 171
validation windows per prefix that are shared with the MiniMax-H3
all-layer experiments. We recompute each metric from the saved per-sample
predictions, including the retrieval gallery, before averaging over
prefixes; the full-set recomputations reproduce the main-table values.
Object persistence still scores only eligible, previously visible targets.
This analysis aligns the validation windows, not the training sets or the
prediction protocols. In particular, MiniMax-H3 direct prediction uses 171
next-observation candidates, whereas the other models use 513 candidates
spanning offsets 1, 2, and 4, so its direct-prediction cell keeps its
dagger; all path-integration entries use 513 waypoint candidates. Values
differ from the main table because both the window set and the gallery
size change, so comparisons are made within each table. The qualitative
picture is stable
under window alignment: the model orderings for depth, camera pose, view
consistency, and direct prediction are identical to those of
\cref{tab:main-results}, instance identity swaps V-JEPA2 and Qwen2.5-VL,
object persistence swaps Qwen2.5-VL and CogVideoX at the bottom of the
ranking, and path integration keeps V-JEPA2 first with only adjacent
swaps elsewhere.

\begin{table}[tbp]
  \centering
  \footnotesize
  \setlength{\tabcolsep}{3pt}
  \caption{\textbf{Spatial readout on the 171 shared validation windows per prefix.}
  Designated layers and prefix aggregation match \cref{tab:main-results}.
  $\dagger$: MiniMax-H3 direct prediction uses only $h=1$; the other
  direct predictors use $h\in\{1,2,4\}$.}
  \label{tab:main-results-common171}
  \begin{tabular}{@{}lrrrrrrr@{}}
    \toprule
    Model (layer) & \makecell{Depth\\loss $\downarrow$}
      & \makecell{Camera\\AUC@30 $\uparrow$}
      & \makecell{Instance\\loss $\downarrow$}
      & \makecell{View\\bal. acc. $\uparrow$}
      & \makecell{Persistence\\error ($^\circ$) $\downarrow$}
      & \makecell{Direct\\R@1 $\uparrow$}
      & \makecell{Path\\R@1 $\uparrow$} \\
    \midrule
    \model{DINOv2-L} (0) & 0.1437 & 0.0421 & 1.4325 & 0.9586 & 66.03 & 0.2875 & 0.1243 \\
    \model{Fast3R} (24) & 0.1371 & 0.3806 & 1.6732 & 0.9817 & 32.21 & 0.0200 & 0.0234 \\
    \model{VideoMAE-L} (23) & 0.3034 & 0.2012 & 4.2136 & 0.9177 & 34.98 & 0.0682 & 0.0838 \\
    \model{V-JEPA2-L} (23) & 0.2802 & 0.1845 & 3.1068 & 0.9225 & 44.38 & 0.0775 & 0.1930 \\
    \model{Qwen2.5-VL} ($-1$) & 0.2270 & 0.0281 & 3.0099 & 0.9384 & 70.28 & 0.2149 & 0.1189 \\
    \model{Wan} (20) & 0.3720 & 0.0653 & 5.1248 & 0.8126 & 60.94 & 0.0331 & 0.0521 \\
    \model{CogVideoX} (20) & 0.4469 & 0.0287 & 6.7726 & 0.7913 & 69.11 & 0.0112 & 0.0244 \\
    \model{Aether} (1) & 0.5399 & 0.0167 & 7.3012 & 0.6916 & 60.31 & 0.0034 & 0.0073 \\
    \model{MiniMax-H3} (30) & 0.4363 & 0.0465 & 6.3714 & 0.8257 & 57.27 & 0.0249$^{\dagger}$ & 0.0117 \\
    \model{LingBot-video} (30) & 0.4432 & 0.0232 & 5.6061 & 0.7765 & 70.03 & 0.0468 & 0.0521 \\
    \bottomrule
  \end{tabular}
\end{table}

\paragraph{Matching the prediction horizon as well as the windows.}
We additionally recompute direct prediction using only the $h=1$ target
from each of the same 171 windows, so that every model is evaluated
against 171 candidates per prefix (\cref{tab:direct-h1-common171}). No
predictor is retrained. This matches the evaluation horizon and the
gallery while retaining the training difference: MiniMax-H3 was trained
on $h=1$ and the other models on $h\in\{1,2,4\}$. The diagnostic therefore
complements the main scores without replacing their protocol. The model
ordering is essentially unchanged ($\rho=0.96$ against the mixed-horizon
column of \cref{tab:main-results-common171}), but the two representations
without a temporal encoder, DINOv2 and Qwen2.5-VL, reach next-observation
R@1 of 0.61 and 0.53, far above every video representation. Because the
next observation is visually close to the last prefix frame, part of this
advantage may reflect feature continuity between adjacent frames rather
than motion-conditioned prediction; the copy baseline of
\cref{sec:probe-validity} would separate the two for these models.

\begin{table}[tbp]
  \centering
  \small
  \caption{\textbf{Direct prediction with a common next-observation target.}
  Mean global R@1 over four prefixes at the designated layer, using the
  same 171 windows and 171 $h=1$ candidates per prefix for every model.
  $\dagger$: next-observation-only training; the other models retain their
  mixed-horizon training.}
  \label{tab:direct-h1-common171}
  \begin{tabular}{@{}lr@{}}
    \toprule
    Model (layer) & $h=1$ R@1 $\uparrow$ \\
    \midrule
    \model{DINOv2-L} (0) & 0.6126 \\
    \model{Fast3R} (24) & 0.0585 \\
    \model{VideoMAE-L} (23) & 0.1930 \\
    \model{V-JEPA2-L} (23) & 0.2281 \\
    \model{Qwen2.5-VL} ($-1$) & 0.5322 \\
    \model{Wan} (20) & 0.1096 \\
    \model{CogVideoX} (20) & 0.0365 \\
    \model{Aether} (1) & 0.0132 \\
    \model{MiniMax-H3} (30) & 0.0249$^{\dagger}$ \\
    \model{LingBot-video} (30) & 0.1418 \\
    \bottomrule
  \end{tabular}
\end{table}

\subsection{Scene-bootstrap intervals}
\label{sec:bootstrap-intervals}
The intervals resample the 69 validation scenes with replacement (1{,}000
draws, seed 0). One draw matrix is shared by every cell, so prefix means,
model differences, prefix effects, and coordinate differences are paired
over the same resampled scenes. Each metric is recomputed from the saved
per-sample predictions as a ratio of per-scene sums, and the recomputed
point estimates equal the reported values for every cell shown. For
retrieval, the gallery stays the full validation set and only the queries
are resampled; the action gains in \cref{tab:c1-action-controls-ci} are
paired per query. The intervals describe evaluation sampling for the
single training seed; they do not estimate training-seed variability.

\begin{table}[tbp]
  \centering
  \caption{\textbf{Main results with 95\% scene-bootstrap intervals.}
  Same cells as \cref{tab:main-results}: designated layer, mean over the evaluated prefixes
  (\cref{eq:main-prefix-average}). Intervals resample the 69 validation scenes (1{,}000 draws, shared by all
  cells, so differences are paired). \textbf{Bold}: the best complete result in each column and every result whose
  paired difference from it has an interval containing zero. $\dagger$: MiniMax-H3 prediction uses 171 of the
  276 validation windows and, for direct prediction, the next-observation target; not ranked.
  }
  \label{tab:main-results-ci}
  \setlength{\tabcolsep}{2.5pt}
  \footnotesize
  \resizebox{\ifdim\width>\linewidth\linewidth\else\width\fi}{!}{%
  \begin{tabular}{@{}lrrrrrrr@{}}
    \toprule
    & \multicolumn{2}{c}{\taskgeometry} & \makecell{Instance\\identity} & \makecell{View\\consistency} & \makecell{Object\\persistence} & \makecell{Direct\\prediction} & \makecell{Path\\integration} \\
    \cmidrule(lr){2-3}\cmidrule(lr){4-4}\cmidrule(lr){5-5}\cmidrule(lr){6-6}\cmidrule(lr){7-7}\cmidrule(lr){8-8}
    Model (layer) & \makecell{Depth\\loss $\downarrow$} & \makecell{AUC@30\\$\uparrow$} & \makecell{Instance\\loss $\downarrow$} & \makecell{Bal.\\acc.\ $\uparrow$} & \makecell{Ang.\ err.\\($^{\circ}$) $\downarrow$} & \makecell{Global\\R@1 $\uparrow$} & \makecell{Global\\R@1 $\uparrow$} \\
    \midrule
    \model{DINOv2-L} (0) & \makecell{\textbf{0.1422}\\{\scriptsize[0.1330, 0.1516]}} & \makecell{0.0412\\{\scriptsize[0.0370, 0.0454]}} & \makecell{\textbf{1.3828}\\{\scriptsize[1.2442, 1.5335]}} & \makecell{0.9557\\{\scriptsize[0.9490, 0.9620]}} & \makecell{65.38\\{\scriptsize[61.56, 69.42]}} & \makecell{\textbf{0.2373}\\{\scriptsize[0.2162, 0.2588]}} & \makecell{0.0900\\{\scriptsize[0.0764, 0.1039]}} \\
    \model{Fast3R} (24) & \makecell{\textbf{0.1386}\\{\scriptsize[0.1254, 0.1527]}} & \makecell{\textbf{0.3870}\\{\scriptsize[0.3558, 0.4170]}} & \makecell{1.6072\\{\scriptsize[1.4446, 1.7828]}} & \makecell{\textbf{0.9808}\\{\scriptsize[0.9760, 0.9847]}} & \makecell{\textbf{32.79}\\{\scriptsize[29.83, 35.98]}} & \makecell{0.0133\\{\scriptsize[0.0103, 0.0169]}} & \makecell{0.0190\\{\scriptsize[0.0148, 0.0242]}} \\
    \model{VideoMAE-L} (23) & \makecell{0.2968\\{\scriptsize[0.2780, 0.3172]}} & \makecell{0.2026\\{\scriptsize[0.1858, 0.2192]}} & \makecell{4.1571\\{\scriptsize[3.7571, 4.5411]}} & \makecell{0.9118\\{\scriptsize[0.9032, 0.9192]}} & \makecell{36.45\\{\scriptsize[33.41, 39.67]}} & \makecell{0.0531\\{\scriptsize[0.0435, 0.0649]}} & \makecell{0.0685\\{\scriptsize[0.0577, 0.0803]}} \\
    \model{V-JEPA2-L} (23) & \makecell{0.2708\\{\scriptsize[0.2529, 0.2888]}} & \makecell{0.1854\\{\scriptsize[0.1696, 0.2026]}} & \makecell{2.9211\\{\scriptsize[2.6350, 3.2244]}} & \makecell{0.9256\\{\scriptsize[0.9183, 0.9323]}} & \makecell{45.31\\{\scriptsize[41.38, 49.32]}} & \makecell{0.0625\\{\scriptsize[0.0537, 0.0716]}} & \makecell{\textbf{0.1588}\\{\scriptsize[0.1449, 0.1739]}} \\
    \model{Qwen2.5-VL} ($-1$) & \makecell{0.2244\\{\scriptsize[0.2104, 0.2387]}} & \makecell{0.0269\\{\scriptsize[0.0244, 0.0295]}} & \makecell{2.9476\\{\scriptsize[2.6158, 3.3016]}} & \makecell{0.9385\\{\scriptsize[0.9310, 0.9460]}} & \makecell{68.59\\{\scriptsize[64.48, 72.11]}} & \makecell{0.1712\\{\scriptsize[0.1564, 0.1875]}} & \makecell{0.0903\\{\scriptsize[0.0776, 0.1033]}} \\
    \model{Wan} (20) & \makecell{0.3609\\{\scriptsize[0.3371, 0.3846]}} & \makecell{0.0650\\{\scriptsize[0.0571, 0.0736]}} & \makecell{4.9301\\{\scriptsize[4.3993, 5.4499]}} & \makecell{0.8215\\{\scriptsize[0.8058, 0.8350]}} & \makecell{62.38\\{\scriptsize[59.15, 65.71]}} & \makecell{0.0260\\{\scriptsize[0.0187, 0.0353]}} & \makecell{0.0405\\{\scriptsize[0.0338, 0.0471]}} \\
    \model{CogVideoX} (20) & \makecell{0.4340\\{\scriptsize[0.4071, 0.4616]}} & \makecell{0.0276\\{\scriptsize[0.0248, 0.0307]}} & \makecell{6.6992\\{\scriptsize[6.0549, 7.3510]}} & \makecell{0.8001\\{\scriptsize[0.7856, 0.8129]}} & \makecell{70.62\\{\scriptsize[66.75, 74.61]}} & \makecell{0.0088\\{\scriptsize[0.0057, 0.0124]}} & \makecell{0.0160\\{\scriptsize[0.0112, 0.0217]}} \\
    \model{Aether} (1) & \makecell{0.5270\\{\scriptsize[0.4938, 0.5601]}} & \makecell{0.0185\\{\scriptsize[0.0163, 0.0210]}} & \makecell{7.1847\\{\scriptsize[6.5741, 7.7540]}} & \makecell{0.6928\\{\scriptsize[0.6808, 0.7049]}} & \makecell{61.18\\{\scriptsize[57.43, 64.58]}} & \makecell{0.0024\\{\scriptsize[0.0009, 0.0039]}} & \makecell{0.0042\\{\scriptsize[0.0021, 0.0066]}} \\
    \model{MiniMax-H3} (30) & \makecell{0.4252\\{\scriptsize[0.3985, 0.4535]}} & \makecell{0.0477\\{\scriptsize[0.0408, 0.0560]}} & \makecell{6.2265\\{\scriptsize[5.5922, 6.8295]}} & \makecell{0.8255\\{\scriptsize[0.8116, 0.8384]}} & \makecell{55.97\\{\scriptsize[52.19, 59.64]}} & \makecell{0.0249$^{\dagger}$\\{\scriptsize[0.0135, 0.0374]}} & \makecell{0.0117$^{\dagger}$\\{\scriptsize[0.0059, 0.0185]}} \\
    \model{LingBot-video} (30) & \makecell{0.4314\\{\scriptsize[0.4042, 0.4590]}} & \makecell{0.0232\\{\scriptsize[0.0206, 0.0262]}} & \makecell{5.5875\\{\scriptsize[4.9998, 6.2688]}} & \makecell{0.7858\\{\scriptsize[0.7696, 0.7994]}} & \makecell{70.30\\{\scriptsize[66.38, 74.08]}} & \makecell{0.0302\\{\scriptsize[0.0239, 0.0380]}} & \makecell{0.0429\\{\scriptsize[0.0356, 0.0510]}} \\
    \bottomrule
  \end{tabular}}
\end{table}

\begin{table}[tbp]
  \centering
  \caption{\textbf{Direct prediction against matched action controls.} Global R@1 at the designated layer,
  mean over prefixes, with 95\% scene-bootstrap intervals. The same frozen probe is evaluated with the true
  action, without the action input, and with a shuffled action; gains are paired per query. Gray: the
  interval of the gain contains zero. Chance is $1/828$ ($1/171$ for MiniMax-H3$^{\dagger}$, next-observation
  target only).}
  \label{tab:c1-action-controls-ci}
  \setlength{\tabcolsep}{2.5pt}
  \footnotesize
  \resizebox{\ifdim\width>\linewidth\linewidth\else\width\fi}{!}{%
  \begin{tabular}{@{}lrrrrr@{}}
    \toprule
    & \multicolumn{3}{c}{Global R@1 $\uparrow$} & \multicolumn{2}{c}{Action gain $\uparrow$} \\
    \cmidrule(lr){2-4}\cmidrule(lr){5-6}
    Model (layer) & \makecell{Full} & \makecell{No action} & \makecell{Shuffled action} & \makecell{Gain vs.\ no action} & \makecell{Gain vs.\ shuffled} \\
    \midrule
    \model{DINOv2-L} (0) & \makecell{0.2373\\{\scriptsize[0.2162, 0.2588]}} & \makecell{0.0223\\{\scriptsize[0.0175, 0.0275]}} & \makecell{0.0691\\{\scriptsize[0.0595, 0.0794]}} & \makecell{0.2150\\{\scriptsize[0.1929, 0.2373]}} & \makecell{0.1682\\{\scriptsize[0.1476, 0.1911]}} \\
    \model{Fast3R} (24) & \makecell{0.0133\\{\scriptsize[0.0103, 0.0169]}} & \makecell{0.0018\\{\scriptsize[0.0003, 0.0033]}} & \makecell{0.0088\\{\scriptsize[0.0060, 0.0121]}} & \makecell{0.0115\\{\scriptsize[0.0082, 0.0151]}} & \makecell{0.0045\\{\scriptsize[0.0015, 0.0075]}} \\
    \model{VideoMAE-L} (23) & \makecell{0.0531\\{\scriptsize[0.0435, 0.0649]}} & \makecell{0.0021\\{\scriptsize[0.0003, 0.0039]}} & \makecell{0.0133\\{\scriptsize[0.0094, 0.0175]}} & \makecell{0.0510\\{\scriptsize[0.0411, 0.0631]}} & \makecell{0.0399\\{\scriptsize[0.0299, 0.0513]}} \\
    \model{V-JEPA2-L} (23) & \makecell{0.0625\\{\scriptsize[0.0537, 0.0716]}} & \makecell{0.0036\\{\scriptsize[0.0015, 0.0057]}} & \makecell{0.0223\\{\scriptsize[0.0166, 0.0287]}} & \makecell{0.0589\\{\scriptsize[0.0498, 0.0679]}} & \makecell{0.0402\\{\scriptsize[0.0323, 0.0483]}} \\
    \model{Qwen2.5-VL} ($-1$) & \makecell{0.1712\\{\scriptsize[0.1564, 0.1875]}} & \makecell{0.0471\\{\scriptsize[0.0396, 0.0556]}} & \makecell{0.0945\\{\scriptsize[0.0839, 0.1054]}} & \makecell{0.1241\\{\scriptsize[0.1093, 0.1395]}} & \makecell{0.0767\\{\scriptsize[0.0643, 0.0912]}} \\
    \model{Wan} (20) & \makecell{0.0260\\{\scriptsize[0.0187, 0.0353]}} & \makecell{0.0082\\{\scriptsize[0.0054, 0.0112]}} & \makecell{0.0187\\{\scriptsize[0.0139, 0.0248]}} & \makecell{0.0178\\{\scriptsize[0.0106, 0.0269]}} & \makecell{0.0072\\{\scriptsize[0.0009, 0.0151]}} \\
    \model{CogVideoX} (20) & \makecell{0.0088\\{\scriptsize[0.0057, 0.0124]}} & \makecell{0.0024\\{\scriptsize[0.0009, 0.0045]}} & \makecell{0.0057\\{\scriptsize[0.0036, 0.0085]}} & \makecell{0.0063\\{\scriptsize[0.0033, 0.0103]}} & \makecell{0.0030\\{\scriptsize[0.0003, 0.0060]}} \\
    \model{Aether} (1) & \makecell{0.0024\\{\scriptsize[0.0009, 0.0039]}} & \makecell{0.0009\\{\scriptsize[0.0000, 0.0021]}} & \makecell{0.0009\\{\scriptsize[0.0000, 0.0018]}} & \textcolor{gray}{\makecell{0.0015\\{\scriptsize[0.0000, 0.0030]}}} & \textcolor{gray}{\makecell{0.0015\\{\scriptsize[-0.0000, 0.0033]}}} \\
    \model{MiniMax-H3} (30) & \makecell{0.0249$^{\dagger}$\\{\scriptsize[0.0135, 0.0374]}} & \makecell{0.0146\\{\scriptsize[0.0057, 0.0249]}} & \makecell{0.0190\\{\scriptsize[0.0099, 0.0294]}} & \makecell{0.0102\\{\scriptsize[0.0029, 0.0195]}} & \textcolor{gray}{\makecell{0.0058\\{\scriptsize[-0.0044, 0.0163]}}} \\
    \model{LingBot-video} (30) & \makecell{0.0302\\{\scriptsize[0.0239, 0.0380]}} & \makecell{0.0103\\{\scriptsize[0.0066, 0.0151]}} & \makecell{0.0190\\{\scriptsize[0.0136, 0.0257]}} & \makecell{0.0199\\{\scriptsize[0.0133, 0.0275]}} & \makecell{0.0112\\{\scriptsize[0.0048, 0.0172]}} \\
    \bottomrule
  \end{tabular}}
\end{table}

\begin{table}[tbp]
  \centering
  \caption{\textbf{Longer versus shorter prefix with 95\% scene-bootstrap intervals.} Designated layer,
  prefix 24 minus prefix 8, oriented so that a positive value means the longer prefix is better (losses and
  errors are negated). Intervals are paired over the same resampled scenes. Gray: the interval contains zero.
  }
  \label{tab:prefix-effects-ci}
  \setlength{\tabcolsep}{2.5pt}
  \footnotesize
  \resizebox{\ifdim\width>\linewidth\linewidth\else\width\fi}{!}{%
  \begin{tabular}{@{}lrrrrrrr@{}}
    \toprule
    Model (layer) & \makecell{Depth\\loss} & \makecell{AUC@30} & \makecell{Instance\\loss} & \makecell{Bal.\\acc.} & \makecell{Ang.\\err.\ ($^{\circ}$)} & \makecell{Direct\\R@1} & \makecell{Path\\R@1} \\
    \midrule
    \model{DINOv2-L} (0) & \makecell{0.0134\\{\scriptsize[0.0084, 0.0188]}} & \makecell{-0.0374\\{\scriptsize[-0.0455, -0.0290]}} & \makecell{-0.2078\\{\scriptsize[-0.3108, -0.1162]}} & \textcolor{gray}{\makecell{0.0009\\{\scriptsize[-0.0093, 0.0125]}}} & \makecell{-15.54\\{\scriptsize[-25.48, -6.25]}} & \textcolor{gray}{\makecell{-0.0205\\{\scriptsize[-0.0568, 0.0133]}}} & \makecell{-0.0543\\{\scriptsize[-0.0845, -0.0229]}} \\
    \model{Fast3R} (24) & \makecell{0.0203\\{\scriptsize[0.0131, 0.0281]}} & \makecell{0.0657\\{\scriptsize[0.0279, 0.1030]}} & \makecell{-0.4441\\{\scriptsize[-0.5860, -0.3110]}} & \textcolor{gray}{\makecell{-0.0021\\{\scriptsize[-0.0083, 0.0030]}}} & \makecell{-12.39\\{\scriptsize[-18.40, -6.44]}} & \textcolor{gray}{\makecell{-0.0036\\{\scriptsize[-0.0145, 0.0072]}}} & \textcolor{gray}{\makecell{-0.0085\\{\scriptsize[-0.0217, 0.0049]}}} \\
    \model{VideoMAE-L} (23) & \makecell{-0.1275\\{\scriptsize[-0.1441, -0.1100]}} & \makecell{-0.1513\\{\scriptsize[-0.1712, -0.1321]}} & \makecell{-8.0407\\{\scriptsize[-9.0544, -6.8937]}} & \makecell{-0.1358\\{\scriptsize[-0.1532, -0.1188]}} & \makecell{-25.25\\{\scriptsize[-32.90, -17.39]}} & \makecell{-0.0664\\{\scriptsize[-0.0894, -0.0447]}} & \makecell{-0.0954\\{\scriptsize[-0.1232, -0.0700]}} \\
    \model{V-JEPA2-L} (23) & \textcolor{gray}{\makecell{-0.0096\\{\scriptsize[-0.0214, 0.0013]}}} & \makecell{-0.0544\\{\scriptsize[-0.0746, -0.0326]}} & \makecell{-1.3906\\{\scriptsize[-1.8088, -0.9595]}} & \textcolor{gray}{\makecell{-0.0104\\{\scriptsize[-0.0242, 0.0028]}}} & \makecell{-31.76\\{\scriptsize[-40.04, -23.32]}} & \textcolor{gray}{\makecell{-0.0048\\{\scriptsize[-0.0266, 0.0169]}}} & \textcolor{gray}{\makecell{-0.0036\\{\scriptsize[-0.0362, 0.0254]}}} \\
    \model{Qwen2.5-VL} ($-1$) & \makecell{0.0196\\{\scriptsize[0.0131, 0.0263]}} & \makecell{-0.0243\\{\scriptsize[-0.0296, -0.0191]}} & \makecell{-1.1563\\{\scriptsize[-1.5449, -0.7944]}} & \textcolor{gray}{\makecell{0.0000\\{\scriptsize[-0.0127, 0.0125]}}} & \makecell{-12.33\\{\scriptsize[-22.91, -3.11]}} & \textcolor{gray}{\makecell{-0.0085\\{\scriptsize[-0.0386, 0.0229]}}} & \makecell{-0.0495\\{\scriptsize[-0.0797, -0.0241]}} \\
    \model{Wan} (20) & \makecell{-0.0490\\{\scriptsize[-0.0636, -0.0345]}} & \makecell{-0.0442\\{\scriptsize[-0.0598, -0.0294]}} & \makecell{-4.1312\\{\scriptsize[-5.0772, -3.2237]}} & \textcolor{gray}{\makecell{-0.0241\\{\scriptsize[-0.0481, 0.0010]}}} & \makecell{-18.42\\{\scriptsize[-28.93, -8.37]}} & \textcolor{gray}{\makecell{-0.0048\\{\scriptsize[-0.0205, 0.0085]}}} & \textcolor{gray}{\makecell{-0.0072\\{\scriptsize[-0.0266, 0.0121]}}} \\
    \model{CogVideoX} (20) & \makecell{-0.0542\\{\scriptsize[-0.0692, -0.0392]}} & \makecell{-0.0300\\{\scriptsize[-0.0390, -0.0223]}} & \makecell{-7.8118\\{\scriptsize[-9.2975, -6.4181]}} & \makecell{-0.0594\\{\scriptsize[-0.0786, -0.0405]}} & \makecell{-16.50\\{\scriptsize[-27.61, -5.93]}} & \textcolor{gray}{\makecell{0.0000\\{\scriptsize[-0.0085, 0.0097]}}} & \textcolor{gray}{\makecell{0.0024\\{\scriptsize[-0.0097, 0.0145]}}} \\
    \model{Aether} (1) & \makecell{-0.0302\\{\scriptsize[-0.0530, -0.0074]}} & \makecell{-0.0171\\{\scriptsize[-0.0214, -0.0129]}} & \makecell{-7.4711\\{\scriptsize[-8.5960, -6.3471]}} & \makecell{0.0246\\{\scriptsize[0.0047, 0.0450]}} & \makecell{-13.51\\{\scriptsize[-22.18, -5.24]}} & \textcolor{gray}{\makecell{0.0000\\{\scriptsize[-0.0048, 0.0048]}}} & \textcolor{gray}{\makecell{0.0000\\{\scriptsize[-0.0072, 0.0072]}}} \\
    \bottomrule
  \end{tabular}}
\end{table}

\begin{table}[tbp]
  \centering
  \caption{\textbf{Generation coordinate differences with 95\% scene-bootstrap intervals.} Block 20, mean
  over prefixes $\{8,12,16,24\}$; each cell is the lower-coordinate score minus the higher-coordinate score
  ($499-999$ for Wan and CogVideoX, $500-1000$ for MiniMax-H3), paired over the same resampled scenes. Gray:
  the interval contains zero.}
  \label{tab:timestep-early-late-ci}
  \setlength{\tabcolsep}{3pt}
  \footnotesize
  \resizebox{\ifdim\width>\linewidth\linewidth\else\width\fi}{!}{%
  \begin{tabular}{@{}lrrrr@{}}
    \toprule
    Model & \makecell{Depth loss $\downarrow$} & \makecell{Instance loss $\downarrow$} & \makecell{View acc.\ $\uparrow$} & \makecell{Localiz.\ err.\ ($^\circ$) $\downarrow$} \\
    \midrule
    \model{Wan} (20) & \makecell{-0.043\\{\scriptsize[-0.050, -0.036]}} & \makecell{1.157\\{\scriptsize[0.647, 1.660]}} & \makecell{0.049\\{\scriptsize[0.040, 0.058]}} & \makecell{-18.8\\{\scriptsize[-22.8, -15.1]}} \\
    \model{CogVideoX} (20) & \makecell{-0.018\\{\scriptsize[-0.024, -0.012]}} & \makecell{-0.537\\{\scriptsize[-1.015, -0.092]}} & \textcolor{gray}{\makecell{0.002\\{\scriptsize[-0.004, 0.009]}}} & \makecell{-6.0\\{\scriptsize[-10.5, -1.7]}} \\
    \model{MiniMax-H3} (20) & \makecell{0.006\\{\scriptsize[0.002, 0.010]}} & \textcolor{gray}{\makecell{0.229\\{\scriptsize[-0.160, 0.613]}}} & \textcolor{gray}{\makecell{0.003\\{\scriptsize[-0.002, 0.008]}}} & \makecell{-4.1\\{\scriptsize[-8.2, -0.3]}} \\
    \bottomrule
  \end{tabular}}
\end{table}

\subsection{Generation coordinate analysis}
\label{sec:generation-coordinate-details}
This separate ScanNet++ extraction study evaluates layers $\{15,20,25\}$
and prefixes $\{8,12,16,24\}$ at scheduler indices $\{499,999\}$ for Wan
and CogVideoX and $\{500,1000\}$ for MiniMax-H3. The main comparison uses
scheduler step 749 for Wan and CogVideoX, which the two indices bracket,
and flow coordinate 0.25 for MiniMax-H3. Coordinates are model-specific
and are not interpreted as a common physical noise level.

\begin{figure}[tbp]
  \centering
  \includegraphics[width=0.95\textwidth]{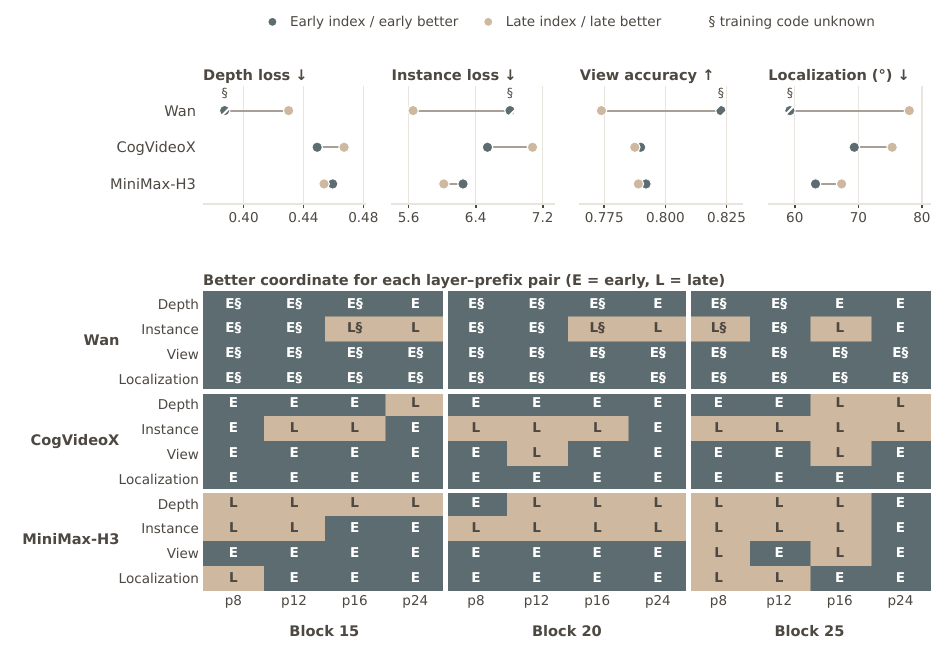}
  \caption{\textbf{Spatial readout across generation coordinates.}
  Top: each model at the early (dark) and the late (light) scheduler index
  at block 20, averaged over prefixes; arrows give the better direction,
  and the values are listed in \cref{tab:timestep-early-late}. Bottom: the
  better coordinate for every layer--prefix pair, E (early) or L (late),
  grouped by block.}
  \label{fig:timestep-analysis}
\end{figure}

\Cref{tab:timestep-early-late} reports final-checkpoint evaluations at
block 20, averaged over all four prefixes, and \cref{fig:timestep-analysis}
shows every layer--prefix pair. Across the twelve layer--prefix pairs per
model and metric, Wan favors the lower index on depth, view consistency,
and object persistence in every pair. CogVideoX favors the lower index on depth in
nine of twelve pairs, on view consistency in ten of twelve, and on
persistence in all twelve. MiniMax-H3 favors the higher index on depth in
ten of twelve pairs and on instance identity in nine of twelve, but the
lower index on view consistency in ten of twelve and on persistence in
nine of twelve. The coordinate effect therefore depends on the model and
the task, with object persistence the one readout that favors the earlier
coordinate for all three models. At block 20, the CogVideoX
view-consistency difference and the MiniMax-H3 instance-identity and
view-consistency differences lie within their paired 95\% intervals
(\cref{tab:timestep-early-late-ci}). The figure and table cover geometry,
instance identity, view consistency, and object persistence; the
direct-prediction and path-integration results in \cref{tab:main-results}
remain measurements at the default extraction coordinates.

\begin{table}[tbp]
  \centering
  \small
  \setlength{\tabcolsep}{3pt}
  \caption{\textbf{Spatial readout at two generation coordinates.}
  Depth, instance-identity loss, view consistency, and object persistence
  at the two scheduler coordinates (block 20, mean over prefixes
  $\{8,12,16,24\}$). Lower/higher coordinates are $499/999$ for Wan and
  CogVideoX and $500/1000$ for MiniMax-H3. Bold marks the better coordinate
  where the paired 95\% interval of the difference excludes zero
  (\cref{tab:timestep-early-late-ci}). At block 20, instance identity
  favors the higher coordinate for Wan and the lower coordinate for
  CogVideoX, while the MiniMax-H3 difference lies within its interval;
  object persistence favors the lower coordinate for all three models.
  These separate-coordinate measurements complement
  the main table, whose generative-model entries use the default
  coordinates.}
  \label{tab:timestep-early-late}
  \begin{tabular}{@{}lcccccccc@{}}
    \toprule
    & \multicolumn{2}{c}{Depth loss $\downarrow$}
    & \multicolumn{2}{c}{Instance loss $\downarrow$}
    & \multicolumn{2}{c}{View acc.\ $\uparrow$}
    & \multicolumn{2}{c}{Localiz.\ err.\ ($^\circ$) $\downarrow$} \\
    \cmidrule(lr){2-3}\cmidrule(lr){4-5}\cmidrule(lr){6-7}\cmidrule(lr){8-9}
    Model & lower & higher & lower & higher & lower & higher & lower & higher \\
    \midrule
    \model{Wan} (20)
      & \best{0.387} & 0.430
      & 6.810 & \best{5.653}
      & \best{0.823} & 0.774
      & \best{59.2} & 78.0 \\
    \model{CogVideoX} (20)
      & \best{0.449} & 0.467
      & \best{6.541} & 7.078
      & 0.790 & 0.787
      & \best{69.4} & 75.3 \\
    \model{MiniMax-H3} (20)
      & 0.459 & \best{0.454}
      & 6.249 & 6.020
      & 0.792 & 0.789
      & \best{63.3} & 67.4 \\
    \bottomrule
  \end{tabular}
\end{table}

\subsection{Motion composition consistency}
\label{sec:se3-diagnostic-details}
We train an external recurrent updater $U$ and readout $D$ only on
one-step camera transitions. Sequential, composed, and swapped updates
start from the same p12 state $s_A$:
\begin{equation}
\begin{aligned}
  \hat z_{\rm seq}&=D(U(U(s_A,T_1),T_2)), &
  \hat z_{\rm dir}&=D(U(s_A,T_2T_1)),\\
  \hat z_{\rm swap}&=D(U(U(s_A,T_2),T_1)), &
  T_1&=T_{AB},\quad T_2=T_{BC}.
\end{aligned}
\label{eq:se3-order-diagnostic}
\end{equation}

The fixed p12 protocol contains 128 training tuples, 32 validation tuples,
and 276 evaluation tuples from 69 scenes. The external recurrent
head is trained only on observed $A\!\to\!B$ transitions, with isolated
endpoint features as targets, using seeds 42--46; it is not either
backbone's native predictor. Cosine endpoint errors are evaluated within
each feature space. Relative gains are
$(e_{\rm baseline}-e_{\rm seq})/e_{\rm baseline}$; the composition
distance is the sequential--direct cosine distance normalized by the
copy-to-endpoint error, using scene-macro means. The normalized order
margin compares swapped-order with correct-order endpoint errors; a
positive value favors the observed order.

All seven evaluated Wan layers and five V-JEPA2 layers improve endpoint
prediction over copying the initial state, and all but Wan block 0 also
improve over the no-action predictor. At the predeclared layers, V-JEPA2
block 11 has a smaller normalized sequential--direct distance than Wan
block 15 (0.0232 versus 0.0377).

\Cref{tab:se3_representative} reports the predeclared representative-layer
comparison, not a best-layer search or an overall model ranking.
Action-only, matched-shuffle, and additive controls do not reveal a
stable order effect, even in strata of high non-commutativity. Identity
padding is only a stress test because the identity lies outside the
training action support. The analytic SE(3) oracle has a positive order
margin of 0.1107\,m, whereas a geometric GRU trained on the same one-step
objective has a sequential RMSE of 0.7010\,m and no reliable order margin.
This limits the attribution of weak latent order effects to the
representation itself.

\begin{table}[tbp]
  \centering
  \caption{\textbf{One-step-to-two-step structural generalization.}
  Predeclared representative layers; differences are Wan block 15 minus
  V-JEPA2 block 11 with paired 95\% intervals. The four comparisons form
  one Holm family. Composition distance and order margin are normalized.}
  \label{tab:se3_representative}
  \footnotesize
  \setlength{\tabcolsep}{4pt}
  \begin{tabular}{@{}lrrrr@{}}
    \toprule
    Metric & Wan (15) & V-JEPA2 (11) & Difference [95\% CI] & Holm $p$ \\
    \midrule
    Gain vs.\ copy $\uparrow$ & 0.2051 & 0.2323 & $-0.0273\ [-0.0814,0.0209]$ & 0.568 \\
    Gain vs.\ no-action $\uparrow$ & 0.0639 & 0.0452 & $0.0187\ [-0.0050,0.0414]$ & 0.384 \\
    Composition distance $\downarrow$ & 0.0377 & 0.0232 & $0.0145\ [0.0107,0.0188]$ & $<0.004$ \\
    Order margin $\uparrow$ & $-0.0001$ & $-0.0007$ & $0.0006\ [-0.0005,0.0019]$ & 0.568 \\
    \bottomrule
  \end{tabular}
\end{table}

\subsection{Action sensitivity of direct future prediction}
\label{sec:direct-action-sensitivity}
We also evaluate each frozen direct predictor on an action set that shares
the same visual prefix. The set contains the relative motions to three
observed endpoints along one trajectory, at offsets 1, 2, and 4. For each
supplied motion, we compare the predicted feature with the three endpoint
features. Within-set matching accuracy is the fraction of cases in which
the highest-similarity endpoint is the one paired with the supplied
motion; chance is $1/3$. The margin is the correct endpoint's cosine
similarity minus the highest similarity among the other two endpoints.

\Cref{tab:direct-action-sensitivity} reports designated-layer means over
four prefixes for the nine models trained on all three offsets. Each
prefix uses 276 action sets with three candidates per set. The results
describe sensitivity to the supplied motion among observed endpoints; they
do not test executed alternative branches or identify a causal
counterfactual effect. The diagnostic also differs from goal-conditioned
action selection: the model predicts a feature given an action, rather
than selecting an action given a goal.

Matching accuracy exceeds chance for every model, but the cosine margins
are small: only DINOv2 (0.029) and V-JEPA2 (0.006) separate the correct
endpoint from the best alternative by a margin clearly above zero, and
the remaining seven margins lie within $\pm0.005$ of zero. The supplied motion
therefore shifts the predicted feature only slightly relative to the
distance between neighboring endpoints. Across the nine models, matching
accuracy ranks the representations much as global retrieval does
($\rho=0.88$ against the direct-prediction column of
\cref{tab:main-results}), so stronger global retrieval coincides with
stronger, though still weak, motion sensitivity.

MiniMax-H3 uses an $h=1$-trained head, so its response to $h=2$ and $h=4$
motions is an additional extrapolation diagnostic. At block 30, its mean
within-set accuracy is 40.74\% over four prefixes, each with 171 action
sets; we report it separately from the mixed-horizon models.

\begin{table}[tbp]
  \centering
  \small
  \caption{\textbf{Motion-dependent endpoint matching.}
  Mean over four prefixes at the designated layer. All listed heads were
  trained on offsets 1, 2, and 4. Matching is within three-candidate sets
  (chance 33.33\%); cosine margins need not be positive even when the mean
  matching accuracy is above chance.}
  \label{tab:direct-action-sensitivity}
  \begin{tabular}{@{}lrr@{}}
    \toprule
    Model (layer) & Matching accuracy (\%) $\uparrow$ & Cosine margin $\uparrow$ \\
    \midrule
    \model{DINOv2-L} (0) & 61.96 & 0.0292 \\
    \model{Fast3R} (24) & 44.57 & $-0.0005$ \\
    \model{VideoMAE-L} (23) & 54.26 & 0.0004 \\
    \model{V-JEPA2-L} (23) & 54.59 & 0.0060 \\
    \model{Qwen2.5-VL} ($-1$) & 47.80 & $-0.0026$ \\
    \model{Wan} (20) & 42.09 & $-0.0010$ \\
    \model{CogVideoX} (20) & 40.31 & $-0.0003$ \\
    \model{Aether} (1) & 35.48 & $-0.0001$ \\
    \model{LingBot-video} (30) & 41.76 & $-0.0049$ \\
    \bottomrule
  \end{tabular}
\end{table}

\subsection{Additional protocol details}
\label{sec:protocol-details}

\paragraph{One-step prediction and history interventions.}
The correct/copy and matched-wrong-motion tests use 276 samples from 69
scenes per prefix; the restricted retrieval pool contains eight
pose-displacement-matched candidates drawn with candidate seed 42. The
p12 object-persistence controls use 165 samples from 65 scenes. Both
studies use probe training seed 42. Reported intervals are paired 95\%
scene-clustered bootstrap intervals. The two tested prediction prefixes
are not strictly target-paired, so agreement across them does not isolate
a history-length effect.

For localization, the appearance query is pooled using a historical
object mask; neither that mask nor the camera pose is supplied to the
spatial readout. Content-only shuffling and reversal retain the final
frame and the temporal identifiers. Moving tokens together with their
identifiers is a permutation sanity check, not a distinct temporal-order
intervention.

\subsection{Supporting controls for predictive spatial readout}
\label{sec:predictive-readout-controls}
The fixed ScanNet++ protocol uses 612 training scenes, 69 held-out
validation scenes, horizons $h\in\{2,4\}$, and three training seeds. The main
question is whether a spatial reader trained on real future states
transfers unchanged to natively predicted features. Two supporting
controls check the specificity and the conditional utility of the spatial
readout.

\paragraph{Target-specific real-feature readout.}
On real target features, Wan's spatial reader reaches scene-macro
normalized-depth MAEs of 0.155 and 0.152 at $h=2$ and $h=4$, compared
with 0.241 and 0.238 for a training-layout prior. Real target features
also outperform within-scene wrong targets and copied context,
demonstrating sample-specific information beyond the layout prior.

\paragraph{Motion-conditioned external prediction.}
For Wan's external predictor, the pooled branch gives a lower downstream
depth MAE under the true motion than under identity motion,
matched-horizon shuffled motion, or an independently trained context-only
predictor. The 0.003--0.007 reductions have paired scene-clustered
intervals excluding zero for all three controls at both horizons. This is
evidence of conditional prediction utility, not of a physical action
intervention.

The same external-predictor analysis on V-JEPA2 gives a normalized-depth
MAE of 0.230/0.228 at $h=2/4$, averaged over three training seeds on the
69 validation scenes. Relative to the independently trained context-only
predictor, the true motion reduces the MAE by 0.0081 [0.0048, 0.0116] and
0.0064 [0.0027, 0.0105], respectively (paired 95\% scene-bootstrap
intervals, 10,000 resamples). The identity-motion and shuffled-motion
controls are also worse at both horizons, with all three seeds agreeing
in direction. These results concern an external predictor and remain
separate from the native-prediction experiment.

\subsection{Action readout protocol and controls}
\label{sec:action-protocol-details}

\paragraph{Goal-conditioned action prediction: candidate-selection setting.}
The candidate experiment uses 844 training queries (650 fit and 194
development) from 289 scenes and 227 validation queries from 77 disjoint
scenes. A shallow scorer receives the frozen current and goal features
plus each candidate 9D action. Ground-truth poses, utilities, realized
endpoints, and scene identities are not given to the scorer. Accuracy is
summarized across three training seeds. The endpoint reference observes
realized outcomes and is diagnostic rather than a candidate-selection
method with the same information.

\paragraph{Goal-conditioned action prediction: continuous-output setting.}
The continuous experiment uses 1,024 training trajectories from 444
scenes (803 fit and 221 development before refitting) and 256 validation
trajectories from 108 disjoint scenes. Epoch selection and normalization
use the development and fit data, respectively. Features are DINOv2's
final projection output, Fast3R block 24, V-JEPA2 block 23, and Wan block
29 at scheduler step 749; the latter differs from Wan's main-table block
20. Separate current and goal projections are combined as
$[c,g,g-c,c\odot g]$; the concatenation control uses the same
optimization budget.

FULL has positive-progress differences of $-3.6$ to $-2.0$ percentage
points relative to CURRENT-ONLY (\cref{fig:goal-conditioning}); only
V-JEPA2's interval excludes zero.
Mean goal reduction is also lower for all four representations, with
intervals excluding zero for DINOv2, V-JEPA2, and Wan. The relational head
improves progress over concatenation for DINOv2 and Fast3R, but only
Fast3R shows a clear goal-reduction gain.

The same-checkpoint intervention substitutes a difficulty-matched goal
from another scene while holding the current state and the learned
weights fixed. Correct-minus-shuffled effects are computed against the
original goal. Five seeds (41--45) are averaged within each sample before
paired scene-bootstrap resampling with 5,000 replicates. This measures
goal sensitivity and utility under the stated input substitution; the
candidate-selection and continuous-output settings remain separate
offline evaluations of the same task.

\subsection{Predictive readout and action prediction metrics}
\label{sec:readout-action-tables}

\Cref{tab:predictive-spatial-interface,tab:continuous-action} provide the
numerical readout and action metrics accompanying
\cref{fig:prediction-action-controls}.

\begin{table}[tbp]
  \centering
  \caption{\textbf{Spatial readout through V-JEPA2's native predictor.}
  Scene-macro normalized-depth MAE $\downarrow$ on 69 validation scenes,
  spatial branch; mean$\pm$SD over three training seeds. The same reader,
  trained on real native target states, is used for real, predicted, and
  copied inputs, except in the explicitly retrained condition. The layout
  prior uses training labels only.}
  \label{tab:predictive-spatial-interface}
  \small
  \setlength{\tabcolsep}{5pt}
  \begin{tabular}{@{}lrr@{}}
    \toprule
    \rowcolor{morandiheader}
    Input / readout condition & $h=2$ & $h=4$ \\
    \midrule
    Real future / original reader      & $0.1824{\pm}0.0003$ & $0.1827{\pm}0.0003$ \\
    Native prediction / frozen reader  & $0.2459{\pm}0.0058$ & $0.2427{\pm}0.0054$ \\
    Native prediction / retrained reader & $0.2233{\pm}0.0008$ & $0.2271{\pm}0.0005$ \\
    Copied context / frozen reader     & $0.2890{\pm}0.0044$ & $0.2974{\pm}0.0041$ \\
    Training layout prior                 & $0.2406$ & $0.2377$ \\
    \bottomrule
  \end{tabular}
\end{table}

\begin{table}[tbp]
  \centering
  \caption{\textbf{Goal-conditioned action prediction: continuous-output setting.}
  FULL uses the relational head; values are mean$\pm$SD over seeds 41--45.
  Progress is the percentage of trajectories with positive progress, and
  goal reduction uses the pose-distance metric defined in the text. The
  final column is the FULL minus CURRENT-ONLY goal reduction with a paired
  95\% scene-bootstrap interval (5,000 replicates); positive favors FULL.
  Seed repeats are averaged within each sample before scene resampling.}
  \label{tab:continuous-action}
  \footnotesize
  \setlength{\tabcolsep}{4pt}
  \begin{tabular}{@{}lrrrr@{}}
    \toprule
    \rowcolor{morandiheader}
    Model & \makecell{Direction ($^{\circ}$)\\$\downarrow$} & \makecell{Progress (\%)\\$\uparrow$} & \makecell{Goal reduction\\$\uparrow$} & \makecell{FULL $-$ CURRENT-ONLY\\goal reduction} \\
    \midrule
    DINOv2-L & $65.3{\pm}1.3$ & $53.8{\pm}1.1$ & $0.008{\pm}0.006$ & $-0.008\ [-0.013,-0.004]$ \\
    Fast3R   & $64.7{\pm}1.3$ & $59.8{\pm}3.4$ & $0.025{\pm}0.007$ & $-0.005\ [-0.012,+0.002]$ \\
    V-JEPA2  & $66.2{\pm}1.8$ & $53.8{\pm}2.4$ & $0.007{\pm}0.007$ & $-0.009\ [-0.015,-0.002]$ \\
    Wan      & $70.0{\pm}1.5$ & $50.2{\pm}3.5$ & $-0.001{\pm}0.009$ & $-0.011\ [-0.017,-0.004]$ \\
    \bottomrule
  \end{tabular}
\end{table}

\subsection{Feature extraction details}
\label{sec:feature-extraction-details}
\Cref{tab:extraction-details} lists the probed layers, the designated
primary layer, and the extraction coordinate of every representation.
\begin{table}[tbp]
  \centering
  \small
  \setlength{\tabcolsep}{5pt}
  \caption{\textbf{Feature extraction details.}
  Layer indices denote the representations used for layer-wise probing.
  For generative models, features are extracted at the specified
  denoising or flow coordinate. The table lists the settings of the main
  sweep; the separate coordinate study additionally uses blocks 15 and
  25.}
  \label{tab:extraction-details}
  \begin{tabularx}{\textwidth}{
    @{}p{0.27\textwidth}
    X
    p{0.16\textwidth}
    p{0.25\textwidth}@{}
  }
    \toprule
    \rowcolor{morandiheader}
    \textbf{Model} &
    \textbf{Probed layers} &
    \textbf{Primary layer} &
    \textbf{Extraction coordinate} \\
    \midrule

    DINOv2-Large
      & final hidden state
      & 0
      & -- \\

    V-JEPA2 ViT-L
      & $\{5,11,17,23\}$
      & 23
      & -- \\

    VideoMAE-Large
      & $\{5,11,17,23\}$
      & 23
      & -- \\

    Fast3R ViT-L
      & $\{24\}$
      & 24
      & -- \\

    Wan2.1-T2V-1.3B
      & $\{0,5,10,15,20,25,29\}$
      & 20
      & scheduler step 749 \\

    CogVideoX-5B-I2V
      & $\{0,5,10,20,30,35,40\}$
      & 20
      & scheduler step 749 \\

    AetherV1
      & $\{1\}$
      & 1
      & scheduler step 749 \\

    MiniMax-H3
      & $\{10,20,30,40,50\}$
      & 30
      & flow coordinate 0.25 \\

    LingBot-video MoE-30B-A3B
      & $\{10,20,30,40,48\}$
      & 30
      & flow coordinate 0.25 \\

    Qwen2.5-VL-7B-Instruct
      & visual merger output
      & $-1$
      & -- \\

    \bottomrule
  \end{tabularx}
\end{table}

\end{document}